\documentclass[12pt]{article}
\usepackage{frExamplee}

\usepackage{graphicx}
\usepackage{apalike}
\usepackage{setspace}
\usepackage{amsmath}
\usepackage{amssymb}
\usepackage{booktabs}
\usepackage[subrefformat=parens]{subcaption}
\usepackage{url}
\usepackage{amsthm}

\usepackage{amsmath}
\usepackage{multirow}
\usepackage{bbm}
\usepackage{float}
\usepackage{fancyhdr}
\usepackage{lastpage}

\usepackage{appendix}

\title{GeoCLR: Georeference Contrastive Learning for Efficient Seafloor Image Interpretation}

\author{
Takaki Yamada, Adam Pr\"{u}gel-Bennett\\
Centre for In Situ and Remote Intelligent Sensing, University of Southampton\\
Southampton SO16 7QF, U.K.\\
\texttt{T.Yamada@soton.ac.uk, apb@ecs.soton.ac.uk} \\
\And
Stefan B. Williams\\
Australian Centre for Field Robotics, The University of Sydney\\
NSW 2006, Australia\\
\texttt{stefan.williams@sydney.edu.au} \\
\AND
Oscar Pizarro\\
Department of Marine Technology, Norwegian University of Science and Technology\\
7491 Trondheim, Norway\\
Australian Centre for Field Robotics, The University of Sydney\\
NSW 2006, Australia\\
\texttt{oscar.pizarro@ntnu.no} \\
\AND
Blair Thornton \\
Centre for In Situ and Remote Intelligent Sensing, University of Southampton\\
Southampton SO16 7QF, U.K.\\
Institute of Industrial Science, The University of Tokyo\\
4-6-1 Komaba Meguro-ku, Tokyo 153-8505, Japan\\
\texttt{B.Thornton@soton.ac.uk} \\
}

\begin{document}

\maketitle

\newpage

\begin{abstract} %
This paper describes Georeference Contrastive Learning of visual Representation (GeoCLR) for efficient training of deep-learning Convolutional Neural Networks (CNNs). The method leverages georeference information by generating a similar image pair using images taken of nearby locations, and contrasting these with an image pair that is far apart. The underlying assumption is that images gathered within a close distance are more likely to have similar visual appearance, where this can be reasonably satisfied in seafloor robotic imaging applications where image footprints are limited to edge lengths of a few metres and are taken so that they overlap along a vehicle’s trajectory, whereas seafloor substrates and habitats have patch sizes that are far larger. A key advantage of this method is that it is self-supervised and does not require any human input for CNN training. The method is computationally efficient, where results can be generated between dives during multi-day AUV missions using computational resources that would be accessible during most oceanic field trials. We apply GeoCLR to habitat classification on a dataset that consists of \url{~}86k images gathered using an Autonomous Underwater Vehicle (AUV).
We demonstrate how the latent representations generated by GeoCLR can be used to efficiently guide human annotation efforts, where the semi-supervised framework improves classification accuracy by an average of 10.2\,\% compared to the state-of-the-art SimCLR using the same CNN and equivalent number of human annotations for training.
\end{abstract}

\newpage
\section{Introduction}
Robotic imaging surveys can enable regional scale understanding of seafloor substrate and habitat distributions. Since visual images are limited to edge lengths of a few metres in water due to the strong attenuation of light, multiple overlapping georeferenced images need to be gathered to describe larger scale patterns that exist on the seafloor. Camera equipped Autonomous Underwater Vehicles (AUVs) achieve this by gathering tens of thousands of images during their dives at close and near-constant altitudes, with  most AUV expeditions lasting many weeks and consisting of several deployments. However, taking advantage of the growing repositories of seafloor images is a challenge because our ability to interpret images cannot keep up with the influx of data. 

Modern machine learning techniques have demonstrated robust, automated image interpretation. Much of the progress in this area has been driven by the availability of generic datasets consisting of over a million human labelled images to supervise the training of deep-learning convolutional neural networks (CNNs). In domains with high learning transferability, this allows deep-learning to be used in applications where the large amount of human effort required to generate labelled training data would be unjustified. However, the appearance of an underwater image is highly sensitive to the environment (e.g.  seawater attenuation properties and turbidity), observation  variables (e.g. range to target) and hardware choices (e.g. lighting and camera configurations), and these factors limit transferability of learning across datasets. This has so far limited large scale generic training datasets from being developed in this domain, and even if these were developed, it is still an open question as to whether these would be as effective as those used in terrestrial applications.

To address this issue, we investigate self-supervision techniques for deep-learning CNNs. Self-supervision is a subset of unsupervised learning, which generates optimised feature descriptors without using human annotations. Self-supervision aims to improve the quality of the image representation by using additional non-image, or image derived data that can be automatically associated with each image to constrain learning. A key advantage of these methods are that they can generate low-dimensional feature vectors on a per dataset basis, making them effective in domains where there is limited transferability of learning across datasets.  Once the representations are obtained, various machine learning techniques such as clustering, content retrieval and few-shot learning can be efficiently applied. Contrastive learning is a form of self-supervision that has demonstrated robust performance gains across many application areas in the image representation learning domain~\cite{chen2020simple,jing2020self,le2020contrastive}. It works by giving similar and dissimilar image pairs to a CNN, optimising the representation that gets generated by mapping similar image pairs within a close distance in the latent representation space, and dissimilar image pairs so that they are separated in this space. To ensure images are similar without relying on human input, most contrastive learning techniques use data augmentation, i.e, applying random transformation to the same image to obtain similar but not identical pairs of images. In this work, we develop a novel method of Georeference enhanced Contrastive Learning for image Representation (GeoCLR), that leverages the 3D location information attached to each image to identify similar pairs of images in target dataset, making the assumption that physically close images are more likely to be similar in appearance than images taken at a larger spatial interval. This assumption is reasonable for the application considered in this work since AUV imagery is taken at close, often overlapping spatial intervals, and when describing substrates and habitats, the features of interest span or recur over spatial scales larger than the footprint of an individual image frame. 

The contributions of this work are:
\begin{itemize}
    \item Development of GeoCLR, a novel contrastive representation learning technique for georeferenced seafloor imagery, leveraging an assumption that a pair of images taken physically close to each other are more likely to have similar appearance than a random pair when identifying similar and dissimilar image pairs in a dataset.
    \item Development of an efficient method to use self-supervised learning outputs to guide human labelling effort to improve the accuracy of low-shot classification. 
    \item Experimental verification of the proposed method's effectiveness through comparison with current state-of-the-art transfer learning and augmentation based contrastive learning, or SimCLR, techniques using a seafloor image dataset consisting of  \url{~}86k AUV images gathered over 12 dives with over \url{~}5k human labels.
\end{itemize}

\section{Background}
\subsection{Representation Learning for Seafloor Imagery}
Visual images of the seafloor contain useful information for mapping substrate and habitat distribution. However, the high-dimensionality and redundant information in raw images is a challenge for classification. Therefore, most algorithmic interpretations first convert images to lower-dimensional representations, or feature spaces, that can be more efficiently analysed. Several types of feature descriptor have been investigated for seafloor image representations~\cite{steinberg2011bayesian,beijbom2012automated,bewley2015hierarchical,kaeli2015online,rao2017multimodal,neettiyath2020}.
In~\cite{beijbom2012automated,neettiyath2020}, colour descriptors are designed to target known targets of scientific interest, such as corals~\cite{beijbom2012automated} and mineral deposits~\cite{neettiyath2020}.
Generic feature descriptors such as Local Binary Patterns (LBP)~\cite{ojala2002multiresolution} and Sparse Coding Spatial Pyramid Matching (ScSPM)~\cite{yang2009linear} have also been applied to capture multi-scale spatially invariant patterns in seafloor images~\cite{bewley2015hierarchical,rao2017multimodal}.
In~\cite{kaeli2015online}, histograms of oriented gradients from image keypoints are applied for clustering and anomaly detection.
These approaches share common steps of selecting effective descriptors and parameter tuning, or feature engineering, that can be time consuming and require knowledge of how targets of interest are expected to appear in the images.

CNNs avoid the need for feature engineering by learning the representations needed to describe the datasets they are trained on. In supervised learning, this is achieved using human annotated examples in a training dataset, where the latent representations and class boundaries to best describe the patterns of interest are simultaneously optimised.
In~\cite{mahmood2018deep}, the deep-learning CNN ResNet~\cite{he2016deep} is trained to classify nine different types of coral in a seafloor image dataset, demonstrating higher classification resolution and accuracy than traditional feature engineering methods. However, the need for large volumes of annotated images limits wide scale use in marine applications as generic training methods and datasets do not currently exist in this domain.

\subsection{Self-supervised learning for Seafloor Imagery}
An alternative approach to train CNNs is self-supervised learning using properties of the data that can be leveraged without the need for direct human supervision. Unlike state-of-the-art supervised training methods such transfer learning, where CNNs are pre-trained with large annotated datasets such as ImageNet~\cite{deng2009imagenet}, self-supervised methods train CNNs on the target dataset itself and so are effective in domains where there is a limited transferability of learning between the target dataset and those available for pre-training. In~\cite{yamada2021learning}, a deep-learning convolutional autoencoder based on AlexNet~\cite{krizhevsky2012imagenet} is used for representation learning of seafloor images.
Autoencoders consist of an encoder and decoder pair, where the encoder maps the original data into low-dimensional latent representations. Next, the decoder reconstructs the original data from the low-dimensional latent representation, where both the encoder and decoder networks are optimised to make the reconstructed data as similar to the original input data as possible. This dimensional reduction attempts to remove redundant information in the raw inputs, retaining only the most important information in compact latent representations at the encoder output. In \cite{yamada2021learning}, the authors developed a Location Guided Autoencoder (LGA) that uses horizontal location information to regularise learning by leveraging the assumption that images captured at nearby locations are more likely to look similar than images that are far apart since seafloor habitats and substrates exhibit patterns larger than the footprint of a single image frame. The method significantly outperformed standard convolutional autoencoders without location regularisation, achieving a factor of 2 improvement in normalised mutual information when applied to clustering and content-based retrieval tasks.
In~\cite{yamada2021leveraging}, the LGA is extended to leverage other types of metadata, such as depth information, where the continuity in measurements have potential correlation with image appearance, where it was demonstrated that these terms can be included without risk of performance degradation through the design of a robust regularisation process.

\subsection{Contrastive Learning Concepts for Image Representation}
The recent development of contrastive learning concepts have demonstrated significant performance gains in self-supervised representation learning~\cite{jing2020self,le2020contrastive}.
The main idea behind contrastive concepts is to simultaneously provide similar and dissimilar image pairs during training, where similar pairs are mapped close to each other in the representation space, and dissimilar pairs are mapped far apart.
These concepts require a binary prior that describes whether the image pairs provided during training are expected to be similar or not.

In~\cite{chen2020simple}, a method to generate similar and dissimilar pairs without any direct human input is developed using data augmentation. The proposed Simple framework for Contrastive Learning of visual Representations (SimCLR) applies random data augmentations to artificially generate similar image pairs, which are then contrasted with dissimilar pairs where different images are used.
The method demonstrated significant gains in performance compared to supervised training using transfer learning approach~\cite{tan2018survey}. 

\section{Contrastive Representation Learning Leveraging Georeference Information}

The use of location information to regularise autoencoder training can enhance the performance of seafloor image representation~\cite{yamada2021learning,yamada2021leveraging}. Here, we investigate whether georeference information can also be leveraged to improve the latent representations generated in contrastive learning~\cite{chen2020simple}. Unlike the modified autoencoder loss functions used in our previous work where location information can be used to loosely regularise learning, the binary similarity condition that is imposed in contrastive learning forces a much stronger constraint on the latent representations that get generated. In order to validate this similarity assumption, we take advantage of the fact that AUVs capture images that often overlap and have footprints that are generally smaller than the patch size of habitats and substrates on the seafloor.

The following subsections give an overview of state-of-the-art modern contrastive learning approaches such as SimCLR~\cite{chen2020simple}, and introduce a novel Georeference Contrastive Learning of visual Representation (GeoCLR) method for efficient representation of spatially contiguous georeferenced imagery.

\begin{figure}[!ht]
     \centering
     \begin{subfigure}[b]{0.45\textwidth}
         \centering
         \includegraphics[width=0.95\textwidth]{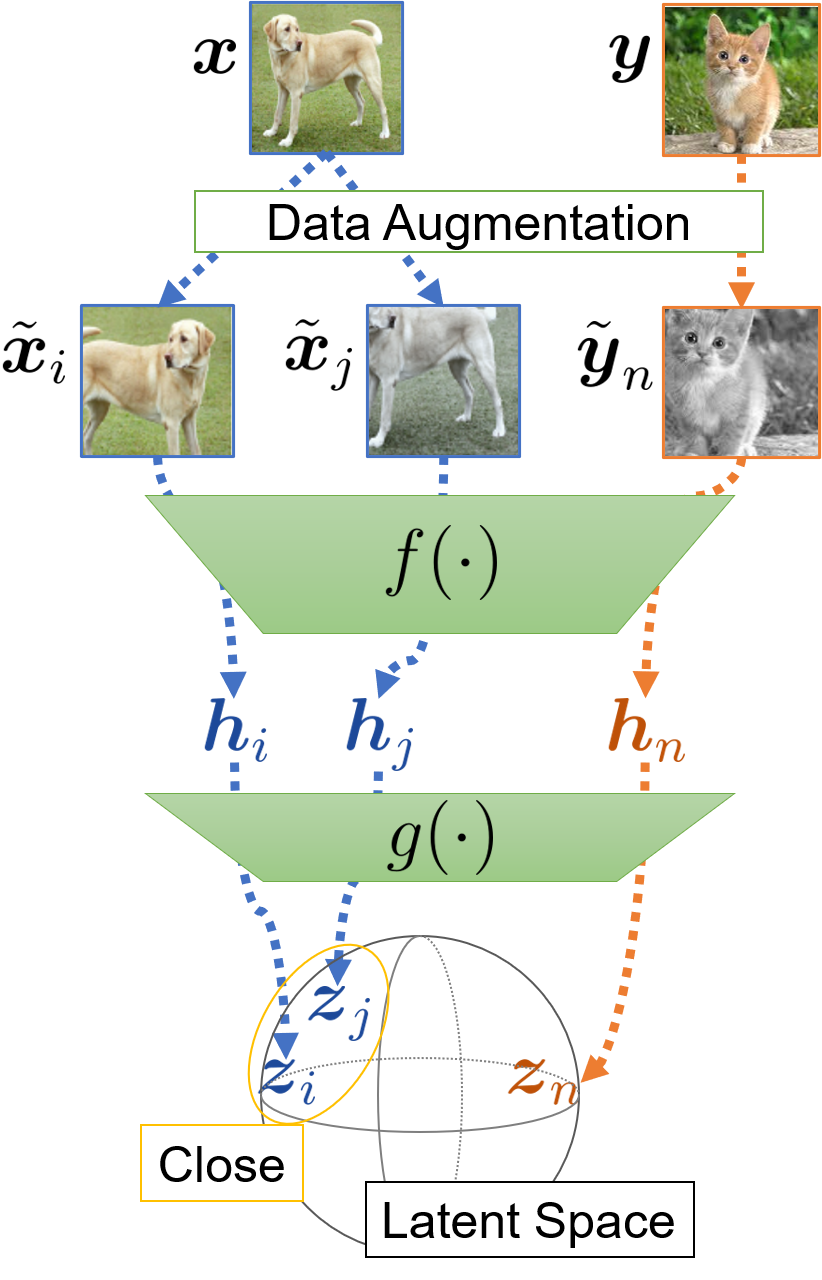}
         \caption{SimCLR}
         \label{fig: simclr overview}
     \end{subfigure}
     \hfill
     \begin{subfigure}[b]{0.45\textwidth}
         \centering
         \includegraphics[width=0.95\textwidth]{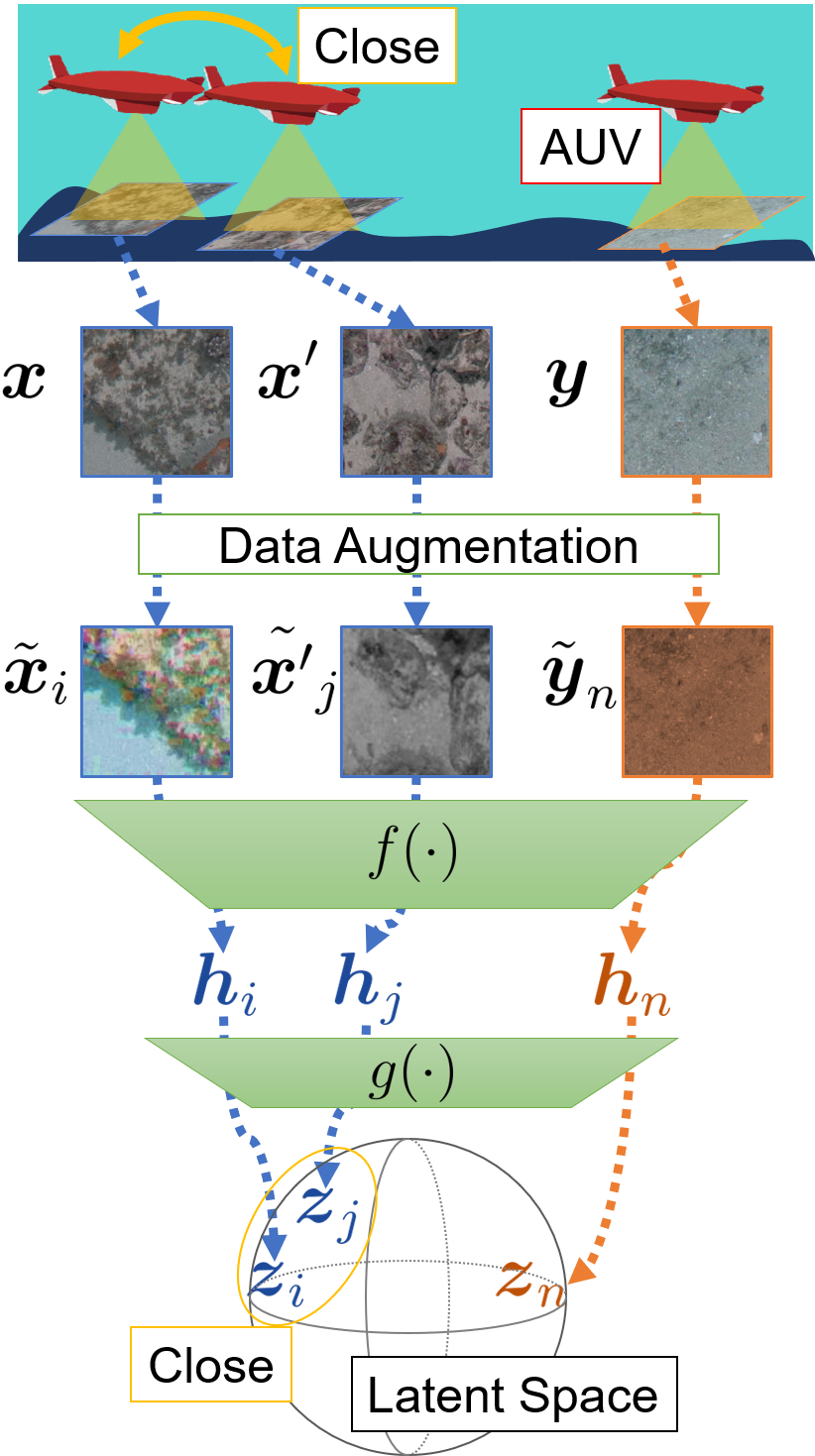}
         \caption{GeoCLR}
         \label{fig: geoclr overview}
     \end{subfigure}
     \caption{Overview of SimCLR and the proposed GeoCLR.
     The two methods apply different conditions to generat similar pairs of images to implement contrastive learning.
     In SimCLR (a), similar image pairs [$\tilde{\boldsymbol{x}}_{i}$, $\tilde{\boldsymbol{x}}_{j}$] are generated by applying different random augmentations to the same image $\boldsymbol{x}$.
     The proposed GeoCLR (b) generates similar pairs [$\tilde{\boldsymbol{x}}_{i}$, $\tilde{\boldsymbol{x'}}_{j}$] using different images that were taken from physically nearby locations, $\boldsymbol{x}$ and $\boldsymbol{x'}$.
     The large range of variability captured in the generated similar pairs allows for robust CNN training.}
\end{figure}

\subsection{SimCLR}
SimCLR learns representations by maximising agreement between differently augmented images generated from the same original image.
The learning framework, illustrated in Figure~\ref{fig: simclr overview}, consists of four parts; data augmentation, base encoder $f(\cdot)$, projection head $g(\cdot)$ and a contrastive loss function. Data augmentation transforms each image $\boldsymbol{x}$ in the target dataset randomly to artificially generate two correlated images, $\tilde{\boldsymbol{x}}_{i}$ and $\tilde{\boldsymbol{x}}_{j}$, where random cropping, colour distortions and Gaussian blur augmentations are applied in this order.
The base encoder $f(\cdot)$ is a CNN that extracts representation vectors from the augmented images.
The method allows any CNN to be used for $f(\cdot)$, where~\cite{chen2020simple} found this approach to be most effective on deeper and wider ResNet~\cite{he2016deep} architectures.
$\boldsymbol{h}_{i} \in \mathbb{R}^{d}$ is a feature vector extracted from $\tilde{\boldsymbol{x}}_{i}$ by the base encoder ($\boldsymbol{h}_{i}=f\left(\tilde{\boldsymbol{x}}_{i}\right)$).
The projection head $g(\cdot)$ is a two layer multilayer perceptron (MLP) to obtain $\boldsymbol{z}_{i} \in \mathbb{R}^{d'}$ ($\boldsymbol{z}_{i}=g(\boldsymbol{h}_{i})$).
The dimension $d'$ of the MLP output are smaller than the dimension $d$ of the base encoder since the contrastive losses defined in lower-dimensional spaces are more efficient for representation learning.
A minibatch of $N$ original images are taken into consideration at each iteration, so $2N$ augmented images including $N$ similar pairs are sampled.
For a similar pair, other $2(N-1)$ augmented images ($\tilde{\boldsymbol{y}}_{n}$ in Figure~\ref{fig: simclr overview}) can be regarded as dissimilar examples within the minibatch.
The Normalised Temperature-scaled Cross Entropy loss function (\emph{NT-Xent})~\cite{sohn2016improved,wu2018unsupervised,oord2018representation} between the similar pair $\tilde{\boldsymbol{x}}_{i}$ and $\tilde{\boldsymbol{x}}_{j}$ is defined as 

\begin{equation} \label{eq. loss term}
\ell_{i, j}=-\log \frac{\exp \left(\operatorname{sim}\left(\boldsymbol{z}_{i}, \boldsymbol{z}_{j}\right) / \tau\right)}{\sum_{k=1}^{2 N} \mathbbm{1}_{[k \neq i]} \exp \left(\operatorname{sim}\left(\boldsymbol{z}_{i}, \boldsymbol{z}_{k}\right) / \tau\right)},
\end{equation}

where sim() denotes cosine similarity, $\mathbbm{1}_{[k \neq i]} \in\{0,1\}$ is the indicator function which is 1 if $k \neq i$, and $\tau$ is the temperature parameter.
The total minibatch loss can be written as,

\begin{equation} \label{eq. loss all}
\mathcal{L}=\frac{1}{2 N} \sum_{k=1}^{N}[\ell(2 k-1,2 k)+\ell(2 k, 2 k-1)].
\end{equation}

The parameters of the base encoder $f(\cdot)$ and the projection head $g(\cdot)$ are updated by a stochastic gradient descent (SGD) optimiser with linear rate scaling~\cite{goyal2017accurate}.

SimCLR can efficiently train CNNs using large unannotated image datasets, where the latent representations derived from the original images $\boldsymbol{x}$ were shown to outperform other state-of-the-art methods in the benchmark classification tasks. It was further shown that fine-tuning of SimCLR trained CNNs can achieve more accurate classification with two orders of magnitude fewer labels than conventional supervised training methods.

\subsection{GeoCLR} \label{sec. geoclr}
A limitation of SimCLR is that the variety of possible image appearances is limited by the types of augmentation used, and only features intrinsic to each image can be efficiently extracted. However, when applied to practical semantic interpretation, we are typically interested in correlating images that show a greater degree of variability than can be described by algorithmic augmentation alone. We predict that the performance of downstream interpretation tasks will benefit if a greater variety of appearances can be integrated into the similar pairs during CNN training. The GeoCLR method proposed in this paper allows great variability to be introduced into the similar image pairs by leveraging the georeference information associated with each image. We argue that the level of variability between images taken nearby will exhibit a level of variability that is more representative of that seen across similar habitats or substrates than augmentation alone.

Figure \ref{fig: geoclr overview} shows the overview of GeoCLR.
In GeoCLR, each similar image pair $[\tilde{\boldsymbol{x}}_{i}$, $\tilde{\boldsymbol{x}}'_{j}]$ is generated from two different images, where $\tilde{\boldsymbol{x}}_{i}$ and $\tilde{\boldsymbol{x}}'_{j}$ are generated from $\boldsymbol{x}'$, which is a different image to $\boldsymbol{x}$ but is taken of a physically nearby location.
For each image $\boldsymbol{x}$ captured at the 3D georeference of $(g_{east},g_{north},g_{depth})$, $\boldsymbol{x'}$ is randomly selected at each iteration from the images which satisfy the following criteria:

\begin{equation} \label{eq. distance criteria}
    \sqrt{(g_{east}'-g_{east})^2 + (g_{north}'-g_{north})^2 + \lambda (g_{depth}'-g_{depth})^2 } \leq r,
\end{equation}

where $(g_{east}',g_{north}',g_{depth}')$ is the 3D georeference of $\boldsymbol{x}'$, $\lambda$ is the scaling factor for depth direction.
Introducing $\lambda > 1$ allows the depth difference between images to be weighted so that the nearby images with large depth gap are not selected, where values of $\lambda < 1$ tend to ignore differences in depth. This flexibility is introduced because the relative impact depth has on image appearance can vary across different application, where for example shallow water application typically have a stronger correlation due to the variable influence of sunlight reaching the seafloor than deep-sea applications.
To identify an image pair, the distance $r$ needs to be larger than the distance between adjacent images taking into account variability in the acquisition interval, and smaller than the patch size of substrates and habitats so that paired images are likely to be similar in appearance. In practise, a small value is advantageous since the similarity assumption is likely to be violated near patch boundaries as $r$ increases. The lower limit for $r$ should also be conservatively set since restricting pairs to only its nearest neighbour means that the same pairing is more likely to be selected multiple times during training, which does not generate any additional information compared to the original SimCLR.

Once $\boldsymbol{x'}$ is selected, the same types of random data augmentation used in SimCLR are applied to each image to obtain the similar pair $[\tilde{\boldsymbol{x}}_{i}$, $\tilde{\boldsymbol{x}}'_{j}]$.
 
\section{Experiment}
The proposed GeoCLR is applied to a dataset consisting of 86,772 georeferenced seafloor images obtained off the coast of Tasmania. The CNN is first trained on all images in the dataset to generate latent representations. Next, classification tasks are given to the trained CNN and the extracted features by the CNN, so that the performance can be evaluated based on classification accuracy. 
A variety of methods are used to train these classifiers, including the use of the latent representations to guide human annotation effort for efficient low-shot semi-supervised learning. The performance of the proposed GeoCLR, is evaluated through comparison with SimCLR and transfer learning under the equivalent conditions.

\subsection{Dataset}

\begin{table}[!ht]
	\centering
	\caption{Tasmania dataset description}
	\label{tab. dataset}
	\begin{tabular}{cc}
		\hline \hline
		Vehicle & Sirius AUV \\
		Camera Resolution & 1,360 $\times$ 1,024 \\
		Camera FoV & 42 $\times$ 34\,deg\\
		Frame Rate & 1\,Hz\\
		Year & 2008 \\
		Location & East Coast of Tasmania, Australia\\ 
		Coordinate & $43.08^{\circ}$S, $147.97^{\circ}$E \\
		Depth & 28 - 96\,m\\
		Altitude & 1.0 - 3.0\,m\\
		Ave. Velocity & 0.5\,m/s\\
		No. of Images & 86,772\\
		No. of Annotations & 5,369\\
		No. of Classes & 6 (See Figure~\ref{fig. class example}) \\
		No. of Dives & 12\\
		\hline \hline
	\end{tabular}
\end{table}

\begin{figure}[!ht]
    \centering
    \includegraphics[width=1.0\linewidth]{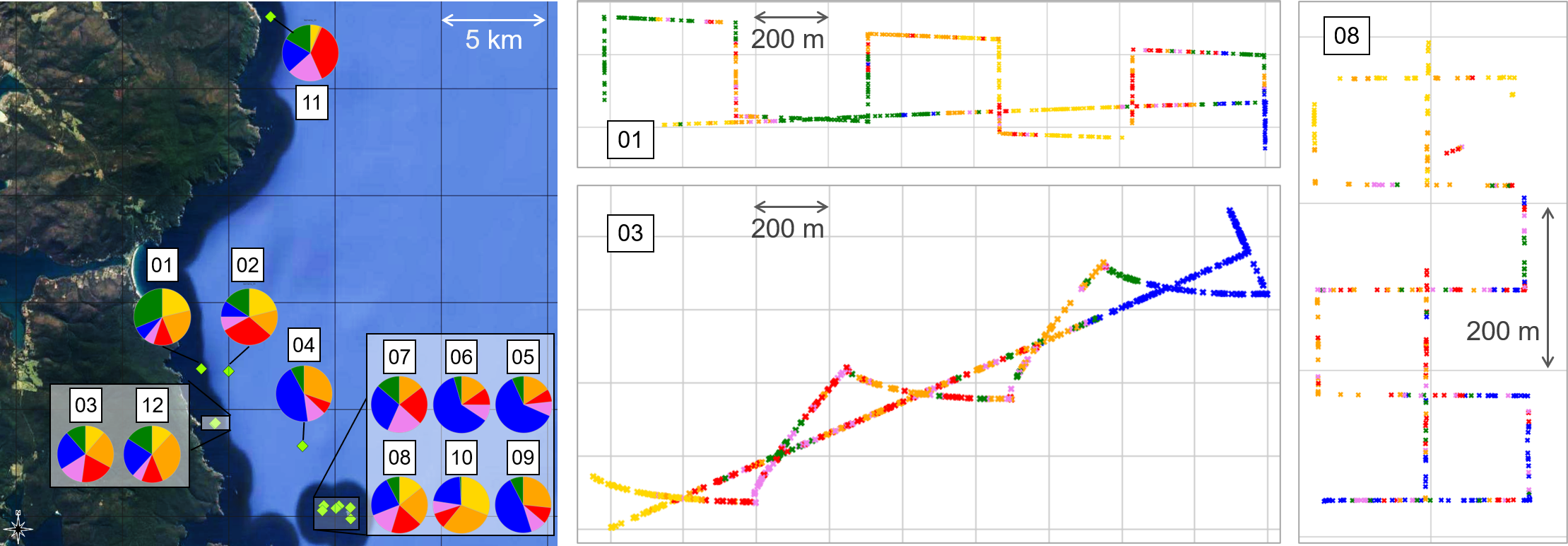}
    \caption{Map of the surveyed area (east coast of Tasmania).
    The images were gathered through 12 AUV deployments.
    The start points of each deployment are shown as green dots. The pie charts show the class distributions according to human expert annotations.
    The same colour scheme is used as in Figure~\ref{fig. class example}, which shows example images of each class.
    The survey paths of Dives 01, 03 and 08 are shown with the human annotated class distributions on the right.
    }
    \label{fig. tasmania map}
\end{figure}

\begin{figure}[!ht]
    \centering
    \resizebox{\textwidth}{!}{
	    \begin{tabular}{ccccccc}
    	     &
            \includegraphics[width=0.18\linewidth]{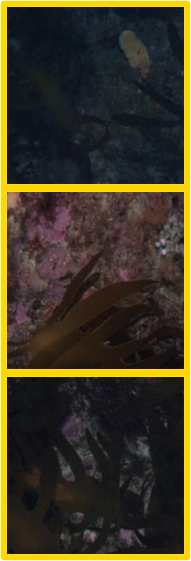}  &
            \includegraphics[width=0.18\linewidth]{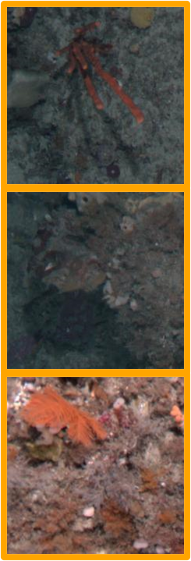}  &
            \includegraphics[width=0.18\linewidth]{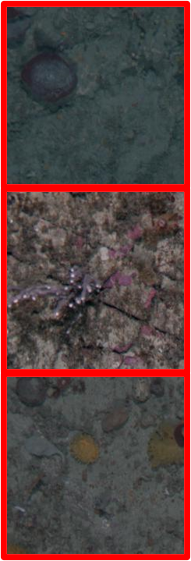}  &
            \includegraphics[width=0.18\linewidth]{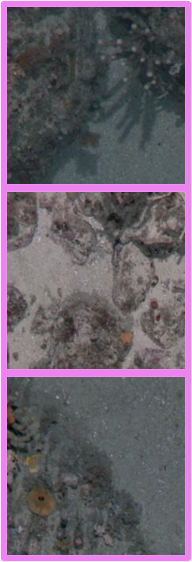}  &
            \includegraphics[width=0.18\linewidth]{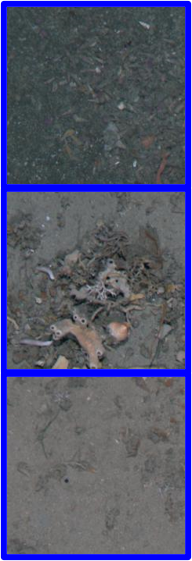}  &
            \includegraphics[width=0.18\linewidth]{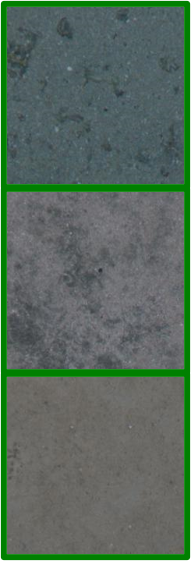} \\
            \textbf{Label} & A & B & C & D & E & F \\
            \textbf{Count} & 531 (9.9\%) & 1,084 (20.2\%) & 903 (16.8 \%) & 598 (11.1 \%) & 1,568 (29.2 \%) & 685 (12.8\%)\\ 
        \end{tabular}
    }
        
    \caption{Class example images together with the number expert human annotations in each class: A - Kelp, B - High Relief Reef, C - Low Relief Reef, D - Reef \& Sand, E - Screw Shell Rubble, F - Sand.}
    \label{fig. class example}
\end{figure}

The Tasmania dataset used for evaluating the proposed GeoCLR~\cite{williams2012} is shown in Figure~\ref{fig. tasmania map} and described in Table~\ref{tab. dataset}.
The dataset consists of 86,772 seafloor images taken by the Australian Centre for Field Robotics' Sirius AUV from a target altitude of 2\,m. The dataset contains habitat and substrate distributions as shown in Figure~\ref{fig. class example}, including kelp (A), a registered essential ocean variable, and rocky reefs (B,C,D), which can form habitats for various conservation targets such as coral and sponges~\cite{GOOS2019}. 5,369 randomly selected images are annotated by human experts into 6 classes, as shown in Figure~\ref{fig. class example}. 50 images randomly selected from each 6 classes (total of 300 images) are used for validation and $M{=}[40,100,200,400,1000]$ images selected from the remaining 5,069 annotated images are used for training, following the evaluation protocol described in subsection~\ref{sec. evaluation protocol}.
The georeference information of each image is determined based on the stereo SLAM pipeline described in~\cite{mahon2008efficient,johnson2010generation}. The original resolution of the images is 1,360 $\times$ 1,024, where the average distance between adjacent images is approximately 0.5 m, so some images partly overlap each other. Prior to analysis, each image in the dataset is re-scaled to a resolution of 2\,mm/pixel based on the imaging altitude. Randomly cropped 224 $\times$ 224 regions of the images are used for training, where validation is performed on the the same sized regions cropped from the centre of the images.

\subsection{GeoCLR Training Configuration}
GeoCLR can by used to train any type of CNN. Here we use the well established ResNet18~\cite{he2016deep} for benchmarking.
The latent representation $\boldsymbol{h}$ and $\boldsymbol{z}$ dimensions are set to $d{=}512$ and $d'{=}128$, respectively.
A minibatch size of $N{=}256$, learning rate of $3.0 \times 10^{-4}$, weight decay of $1.0 \times 10^{-4}$, temperature $\tau{=}0.07$ in equation(\ref{eq. loss term}) was used for all experiments. 
The threshold of closeness and depth scaling factor in equation~\ref{eq. distance criteria} were set to $r{=}1.0\,m$ and $\lambda{=}1.0$, respectively. The value for $r$ is conservative compared to the expected substrate and habitat patch size in the surveyed region, and was chosen to yield 2 to 4 nearby images based on the average distance between images (see Table~\ref{tab. dataset}). This minimises the probability of non-similar image pairs being selected near patch boundaries and the likelihood of duplicate pairs being selected during training. Results on the sensitivity of learning performance to a range of $r{=}0, 1.0, 3.0, 5.0, 10.0$\,m included in the Appendix demonstrate this trend, with no benefit seen beyond $r=3.0$\,m for the dataset analysed in this paper. As expected, a small value of $r$ is advantageous, with the limit of sufficiently nearby images being unavailable if it is set too small. The mean range between best and worst performing conditions is 3.7\,\% for equivalent $M$ and $\lambda$ numbers. The value of $\lambda$ was chosen to evenly treat horizontal and vertical displacement between images. Analysis in the Appendix for $\lambda{=}0, 0.5, 1.0, 3.0, 5.0, 10.0$ shows that the performance sensitivity small. The mean range between best and worst performing conditions is 1.7\,\% for equivalent $M$ and $r$ numbers. The sensitivity and optimum value for $\lambda$ is likely to be dataset dependent, and more significant in datasets that have high rugosity.

Other than the method for generating similar image pairs, identical parameters were used for GeoCLR and SimCLR to allow for comparison. Both method are trained on the all 86,772 images in the dataset. We also benchmark the performance of the proposed method against conventional supervised transfer learning using ResNet18 that was pre-trained on ImageNet.

Though deeper CNN architectures, larger minibatch sizes and epoch are known to provide accuracy gains for SimCLR, these above parameters are set considering the computational power that can be reasonably deployed in the field, where access to high-performance computers networks is limited. The workstation used for experiments in this paper used a single NVIDIA TITAN RTX with 24 GB VRAM. The GeoCLR training and fine tuning with pseudo-labelling carried out in this work each took approximately a day (26 hours for GeoCLR training and a few minutes for fine-tuning) for the dataset of \url{~}86k images gathered in 24 hours of bottom time over multiple AUV dives. This indicates that the results could be made available in timeframes relevant to assist planning and interpretation between dives during multi-day field expeditions.
 
\subsection{Evaluation Protocol}\label{sec. evaluation protocol}
CNNs trained using three different approaches (ImageNet, SimCLR, GeoCLR) are evaluated following the protocol used by~\cite{chen2020simple}. Once the CNNs are trained, the latent representations they generated are analysed using different classifiers; a linear logistic regression, a non-linear Support Vector Machine with a Radial Basis Function kernel (SVM with RBF), and a fine-tuned CNN classifier. The logistic regression and SVM with RBF are both trained on the latent representation space output $\boldsymbol{h}$ of ResNet18 after CNN training. For fine-tuning, a minibatch size of 256, Adam optimiser with learning rate of $3.0 \times 10^{-4}$ and no weight decay. The macro averaged f$_1$ score over 6 classes determined from the independent validation set is used to compare the classification accuracy of each training method. All experiments are repeated ten times in each configuration, where the standard deviation (SD) of scores is shown alongside the mean value to describe variability.

We perform experiments to evaluate classification performance using $M{=}[40,100,200,400,1000]$ training examples selected using the following sampling strategies: \begin{itemize}
    \item \emph{Balanced}: $M$ annotated images are selected so that all classes are equally represented.
    \item \emph{Random}: $M$ annotated images are randomly selected without any constraint.
    \item \emph{H-$k$means}: $M$ annotated images are selected using hierarchical $k$means to evenly represent different regions of the latent representation space learned through self-supervised training. 
\end{itemize}

Class-balanced training examples can be considered ideal for supervised learning in applications where all classes are of equal importance. However, this requires significant human effort to determine the relevant classes and identify images corresponding to each class, which is not practical for most field survey scenarios. The \emph{random} method is relevant for most survey scenarios, and can make the effort in annotation more manageable in situations where the class distribution in the target dataset is not known. However, the method suffers when the number of images in each class is not balanced, since classes are represented in proportion to their relative abundance, those with small populations tend to exhibit poor performance. The hierarchical $k$means clustering~\cite{nister2006scalable}, or \emph{H-$k$means}, method allows for balanced representation of the variety of images present in a dataset without the need for additional human effort, and was shown to be effective for guiding human labelling effort in~\cite{yamada2022guiding}. In this method, $k$means clustering is first applied to latent representations with $k{=}m$ to find representative clusters of images in the dataset. An appropriate value for $m$ can be automatically determined for each dataset using the elbow method~\cite{satopaa2011finding}, where a value of $m{=}10$ was found to be appropriate for the dataset used in this experiment. Next, each cluster is further subdivided using $k$means clustering where $k{=}M/m$, to identify images closest to the centroid of each subdivided cluster, which is prioritised for human annotation. If the latent representations describe original images appropriately, the cluster boundaries found in the first $k$means clustering are expected to approximate the class boundaries, allowing the class imbalance problem in \emph{random} selection to be eased by selecting the same number of images from each cluster. 
The second $k$means clustering avoids selecting similar samples from within each cluster, so that the full variety of images in the dataset can be represented by a small number of annotations.
This \emph{H-$k$means} selection was shown to outperform \emph{random} selection when appropriate latent representations are generated~\cite{yamada2022guiding}. The same work also demonstrated the use of pseudo-labels, generated from the predictions of classical classifiers applied to the latent representations, for CNN fine-tuning, which is also examined in this work.

Two sets of experiments are performed. First, we assess the performance of CNNs using the \emph{balanced} selection strategy. Although this is not realistic for practical field scenarios, the results represent the expected upper bound of performances, and allows the fundamental performance of the three latent representation learning strategies (ImageNet, SimCLR, GeoCLR) to be compared. The second set of experiments compares the performance using \emph{random} and 
\emph{H-$k$means} based selection strategies, including the use of pseudo labelling with linear logistic regression (PL-linear), a non-linear SVM with RBF (PL-SVM) for CNN fine-tuning. The sampling strategies can both realistically be implemented in field survey scenarios since they do not assume any prior knowledge of the datasets, and the images that require annotation can be rapidly identified in a fully unsupervised manner.

\subsection{Result}
\subsubsection{Class-balanced training evaluation}

\begin{table}[!t]
	\centering
	\caption{CNN training method comparison on class balanced training subset}
	\label{tab. cnn training comparison}
	\centering
	\resizebox{\textwidth}{!}{
		
		\begin{tabular}{ccc|cccccc}
			\hline \hline
			\multirow{2}{*}{\textbf{\begin{tabular}[c]{@{}c@{}}Config. \\ Label\end{tabular}}} &
			\multirow{2}{*}{\textbf{\begin{tabular}[c]{@{}c@{}}CNN \\ Training\end{tabular}}} &
			\multirow{2}{*}{\textbf{Classifier}} &
			\multicolumn{5}{c}{\textbf{Number of Annotations ($M$)}} \\
			& & & 40 & 100 & 200 & 400 & 1000            \\
			\hline
			A1 &  ImageNet & linear  &  54.9$\pm$4.7 & 61.6$\pm$2.8 & 63.0$\pm$2.2 & 67.5$\pm$2.2 & 67.4$\pm$2.1\\
			A2 &  ImageNet & SVM  &  47.0$\pm$4.9 & 55.3$\pm$4.9 & 60.2$\pm$2.3 & 66.2$\pm$1.1 & 69.7$\pm$1.1\\
			A3 &  ImageNet & Res18 & 58.9$\pm$2.6 & 65.5$\pm$2.7 & 68.2$\pm$2.5 & 71.2$\pm$1.7 & 73.8$\pm$1.3\\
			\hline
			B1 &  SimCLR & linear  &  62.5$\pm$2.7 & 65.2$\pm$2.8 & 67.1$\pm$1.2 & 69.2$\pm$2.2 & 71.8$\pm$1.0\\
			B2 &  SimCLR & SVM  &  62.4$\pm$2.7 & 66.9$\pm$1.8 & 69.2$\pm$1.8 & 71.8$\pm$1.4 & 74.1$\pm$1.0\\
			B3 &  SimCLR & Res18 & 53.4$\pm$4.4 & 61.3$\pm$2.2 & 65.5$\pm$2.0 & 68.9$\pm$2.7 & 72.4$\pm$0.9\\
			\hline
			C1 &  GeoCLR & linear & \textbf{63.8}$\pm$2.9 & 67.8$\pm$2.4 & 71.4$\pm$1.4 & 72.9$\pm$1.8 & 74.9$\pm$1.0\\
			C2 &  GeoCLR & SVM & 61.7$\pm$2.5 & \textbf{70.1}$\pm$2.4 & \textbf{74.5}$\pm$1.4 & \textbf{75.8}$\pm$1.4 & \textbf{78.3}$\pm$1.1\\
			C3 &  GeoCLR & Res18 & 53.6$\pm$5.3 & 62.8$\pm$2.2 & 66.2$\pm$2.9 & 69.5$\pm$1.9 & 73.2$\pm$1.3\\
			\hline \hline
		\end{tabular}
	}
	\begin{flushleft}
		The CNNs are trained using three different method (Supervised Learning by ImageNet, SimCLR and GeoCLR).
		The latent representations ($\boldsymbol{h}$) extracted from the $M$ annotated images by each CNN are used for logistic regression classification (linear) and SVM (with RBF) training.
		Also the CNNs are fine-tuned on the same subsets of images.
		The $M$ images are selected so that all 6 classes in the dataset are evenly described.
		The classifiers are trained 10 times with different random seed, and mean and SD values of f$_{1}$ scores (macro averaged) are shown.
		The best score for each $M$ is shown as bold.
	\end{flushleft}	
\end{table}

Table~\ref{tab. cnn training comparison} shows the macro averaged f$_1$ scores of each CNN training and classifier configuration on the class-balanced subsets of the annotated images.
The results show that GeoCLR has the best performance for all values of $M$, with the linear classifier (C1) showing the best performance for $M{=}40$, and the SVM classifier (C2) best for all other $M$ values. The latent representations generated using GeoCLR achieves an average 7.4\,\% and 5.2\,\% increase in performance compared to the best performing ImageNet and SimCLR trained configurations.

Among the ImageNet pre-trained CNN (A$\ast$), the CNN fine-tuned using $M$ images (A3) achieves the highest accuracy for all $M$ with an average performance gain of 6.6\,\%. This is owed to the capacity of CNNs to simultaneously optimise feature extraction and classification during training. In A1 and A2, the lower level feature extractor optimised on ImageNet is not updated. The inferior performance compared to A3 indicates that the latent representations generated using ImageNet are suboptimal for the seafloor images used in this work, failing to describe their useful distinguishing features.

In contrast, for SimCLR and GeoCLR trained CNNs (B$\ast$ and C$\ast$) the fine-tuned scores for the ResNet18 classifier are lower than the scores of linear and SVM classifier.
This shows that the constraining effect of contrastive learning in SimCLR or GeoCLR training generates highly optimised latent representations. Since conventional fine-tuning does not maintain this constraining effect, it degrades performance, achieving a similar level of accuracy as fine-tuning of ImageNet pre-trained CNN in A3 for larger values if $M$.
This finding is in contrast to the results of~\cite{chen2020simple}, where fine-tuning of CNNs trained using SimCLR significantly outperform linear classifiers applied to latent representation space for generic terrestrial image datasets analysed. Possible reasons for the difference in behaviour is the relatively high dimensionality of $\boldsymbol{h}$ ($d{=}512$) compared to the small number of classes (6) in the dataset considered in this paper, combined with the continuous transition of image appearance across the class boundaries, both of which are different to terrestrial benchmark datasets, which typically have a larger number of classes with discrete boundaries, both of which can make the latent representation more sensitive to the constraining effect of contrastive learning. 

Figure~\ref{fig. acc plot balanced} shows representative configurations from Table~\ref{tab. cnn training comparison}.
The proposed GeoCLR with a SVM classifier (C2) outperforms all other configurations except for B2 when $M{=}40$. Having said this, the best performance for $M{=}40$ is achieved by the GeoCLR with a linear classifier (C1) as can be seen in Table~\ref{tab. cnn training comparison}.
When the CNNs are fine-tuned, transfer learning with ImageNet (A3) outperforms fine-tuned SimCLR and GeoCLR (B3 and C3).

\subsubsection{Data selection method comparison}

\begin{table}[!ht]
	\centering
	\caption{Data selection method comparison}
	\label{tab. data selection comparison}
	\centering
	\resizebox{\textwidth}{!}{
		
		\begin{tabular}{cccc|cccccc}
			\hline \hline
			\multirow{2}{*}{\textbf{\begin{tabular}[c]{@{}c@{}}Config. \\ Label\end{tabular}}} &
			\multirow{2}{*}{\textbf{\begin{tabular}[c]{@{}c@{}}CNN   \\ Training\end{tabular}}} &
			\multirow{2}{*}{\textbf{Classifier}} &
			\multirow{2}{*}{\textbf{\begin{tabular}[c]{@{}c@{}}Data\\  Selection\end{tabular}}} &
			\multicolumn{5}{c}{\textbf{Number of Annotations ($M$)}} \\
			& & &  & 40 & 100 & 200 & 400 & 1000            \\
			
			\hline
			D1 &  ImageNet & linear & random & 45.4$\pm$5.8 & 57.0$\pm$3.2 & 61.3$\pm$2.8 & 63.3$\pm$2.9 & 67.5$\pm$2.2\\
			D2 &  ImageNet & linear & H-$k$means & 49.5$\pm$5.6 & 58.0$\pm$4.7 & 64.4$\pm$3.4 & 66.5$\pm$2.6 & 68.8$\pm$1.8\\
			D3 &  ImageNet & SVM & random & 35.5$\pm$4.4 & 50.2$\pm$3.7 & 58.4$\pm$3.3 & 63.7$\pm$1.4 & 67.9$\pm$1.0\\
			D4 &  ImageNet & SVM & H-$k$means & 43.0$\pm$3.4 & 57.5$\pm$2.8 & 63.7$\pm$1.0 & 67.0$\pm$1.4 & 69.7$\pm$1.3\\
			
			D5 &  ImageNet & Res18 & random & 55.5$\pm$3.1 & 63.2$\pm$3.2 & 67.0$\pm$2.0 & 69.7$\pm$2.4 & 72.8$\pm$2.3\\
			D6 &  ImageNet & Res18 & H-$k$means & 51.8$\pm$5.7 & 64.1$\pm$2.0 & 67.6$\pm$2.7 & 73.1$\pm$1.3 & 74.1$\pm$1.7\\
			D7 &  ImageNet & Res18 & PL-linear & 51.9$\pm$5.7 & 62.2$\pm$4.5 & 68.6$\pm$1.8 & 70.9$\pm$2.0 & 71.9$\pm$2.3\\
			D8 &  ImageNet & Res18 & PL-SVM & 46.4$\pm$5.4 & 58.9$\pm$3.4 & 67.1$\pm$1.6 & 69.9$\pm$1.7 & 72.6$\pm$2.1\\
			
			\hline
			E1 &  SimCLR & linear & random & 55.0$\pm$4.0 & 63.8$\pm$3.0 & 66.3$\pm$2.1 & 67.7$\pm$3.2 & 71.2$\pm$1.1\\
			E2 &  SimCLR & linear & H-$k$means & 61.9$\pm$2.6 & 66.5$\pm$1.6 & 68.2$\pm$1.4 & 69.3$\pm$2.7 & 69.5$\pm$1.9\\
			E3 &  SimCLR & SVM & random & 47.2$\pm$5.5 & 64.5$\pm$2.4 & 68.8$\pm$1.6 & 72.0$\pm$2.2 & 73.5$\pm$0.7\\
			E4 &  SimCLR & SVM & H-$k$means & 58.0$\pm$2.0 & 67.6$\pm$1.5 & 70.9$\pm$1.3 & 71.7$\pm$1.8 & 73.7$\pm$1.5\\
			
			E5 &  SimCLR & Res18 & random & 49.5$\pm$7.5 & 60.3$\pm$2.2 & 64.5$\pm$2.4 & 67.7$\pm$1.8 & 71.5$\pm$2.3\\
			E6 &  SimCLR & Res18 & H-$k$means & 56.4$\pm$3.3 & 65.2$\pm$2.2 & 66.2$\pm$1.7 & 69.3$\pm$2.0 & 70.1$\pm$1.2\\
			E7 &  SimCLR & Res18 & PL-linear & 64.3$\pm$2.2 & 68.8$\pm$1.4 & 69.5$\pm$1.7 & 70.5$\pm$1.7 & 72.8$\pm$1.3\\
			E8 &  SimCLR & Res18 & PL-SVM & 63.1$\pm$2.8 & 69.8$\pm$1.8 & 70.6$\pm$1.1 & 72.7$\pm$1.1 & 72.9$\pm$0.8\\
			
			\hline
			F1 &  GeoCLR & linear & random & 58.9$\pm$5.0 & 67.8$\pm$2.7 & 70.8$\pm$1.7 & 72.7$\pm$2.5 & 75.1$\pm$1.4\\
			F2 &  GeoCLR & linear & H-$k$means & \textbf{65.8}$\pm$2.9 & 70.5$\pm$1.7 & 72.8$\pm$2.0 & 73.0$\pm$2.1 & 74.6$\pm$2.5\\
			F3 &  GeoCLR & SVM & random & 53.2$\pm$5.9 & 68.8$\pm$3.1 & 72.9$\pm$2.2 & 75.5$\pm$1.0 & 77.5$\pm$1.2\\
			F4 &  GeoCLR & SVM & H-$k$means & 55.3$\pm$4.2 & 71.8$\pm$1.6 & \textbf{74.6}$\pm$1.5 & \textbf{76.6}$\pm$1.2 & \textbf{79.0}$\pm$1.0\\
			
			F5 &  GeoCLR & Res18 & random & 49.5$\pm$7.9 & 60.3$\pm$3.8 & 65.2$\pm$1.7 & 69.0$\pm$3.0 & 73.2$\pm$1.9\\
			F6 &  GeoCLR & Res18 & H-$k$means & 56.5$\pm$3.4 & 65.5$\pm$1.4 & 66.8$\pm$1.9 & 70.9$\pm$1.3 & 73.9$\pm$1.7\\
			F7 &  GeoCLR & Res18 & PL-linear & 64.2$\pm$2.5 & 71.7$\pm$2.3 & 72.7$\pm$1.6 & 73.5$\pm$1.4 & 75.7$\pm$1.6\\
			F8 &  GeoCLR & Res18 & PL-SVM & 63.7$\pm$2.1 & \textbf{72.0}$\pm$2.1 & 72.5$\pm$1.5 & 74.3$\pm$1.0 & 75.2$\pm$1.3\\
			\hline \hline
		\end{tabular}
	}
	\begin{flushleft}
		The same CNNs as Table~\ref{tab. cnn training comparison} where different data selection strategies (\emph{random} and  \emph{$H$-$k$means} are used in the downstream classification task. In contrast to the \emph{balanced} selection strategy shown in Table~\ref{tab. cnn training comparison}, these selection strategies do not require prior analysis by humans and so are available for analysis of data as it get collected in the field. The same classifiers (linear, SVM, fine-tuned ResNet18) are investigated.
		For fine-tuning the CNNs, pseudo-labels generated by linear classifier ($\ast$-7) or SVM ($\ast$-8) are used.
		The classifiers are trained 10 times with different random seed, and mean and SD values of f$_{1}$ scores (macro averaged) are shown.
		The best score for each $M$ is shown as bold.
	\end{flushleft}
\end{table}

The performance using the \emph{random} and \emph{H-$k$means} training data selection strategies, both of which do not need prior human input to understand the datasets, are shown in Table~\ref{tab. data selection comparison} for different values of $M$.

The different CNN training methods shows the same trend as the previous results with \emph{balanced} training data selection (Table~\ref{tab. cnn training comparison}). When the classifiers are trained on the latent representations, GeoCLR outperforms SimCLR and ImageNet pre-training, achieving and average performance gains of 6.3\,\% and 20.0\,\%, respectively across all $M$. As previously observed, fine-tuning SimCLR and GeoCLR trained CNNs degrades their performance. However, the pseudo-labelling introduced in (E7, E8, F7, F8) mitigates this effect by using a larger number of images for fine-tuning, which avoids the problem of overfitting that can occur when only a small number of images are used in fine tuning. This effect is strongest for small values of $M{=}40,100$, where performance gains of 13.1\,\% and 8.0\,\% are achieved for both GeoCLR and SimCLR compared to equivalent configurations that do not use PL.

\begin{figure}[!t]
     \centering
     \begin{subfigure}[b]{0.5\textwidth}
         \centering
         \includegraphics[width=\textwidth]{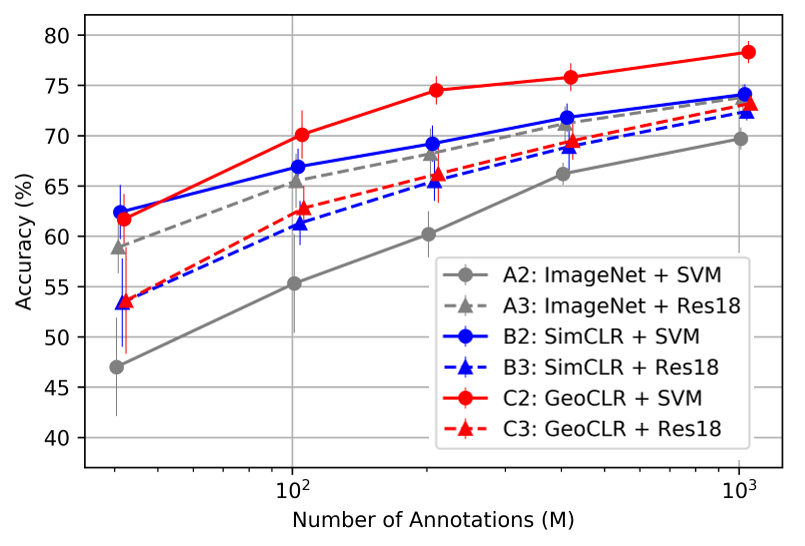}
         \caption{CNN training method comparison for class-balanced training (Table~\ref{tab. cnn training comparison}) }
         \label{fig. acc plot balanced}
     \end{subfigure}\\
     \vspace{10 mm}
     \centering
     \begin{subfigure}[b]{0.5\textwidth}
         \centering
         \includegraphics[width=\textwidth]{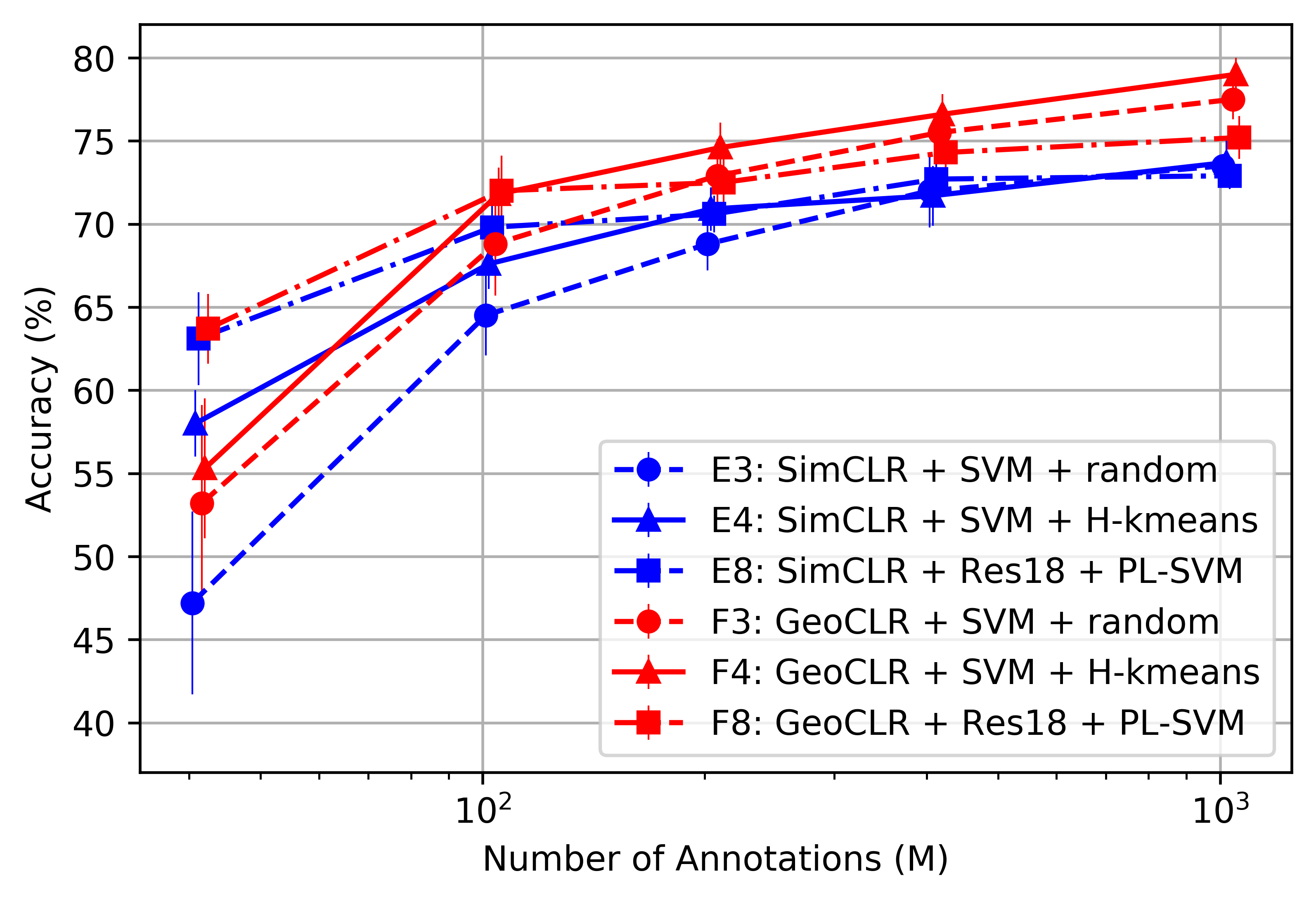}
         \caption{Data selection method comparison for SimCLR and GeoCLR (Table~\ref{tab. data selection comparison})}
         \label{fig. acc plot random hk}
     \end{subfigure}
     \caption{Representative Configurations from (a) Table~\ref{tab. cnn training comparison} and (b) Table~\ref{tab. data selection comparison}.
     (a) When the CNNs are trained on the class-balanced subsets,      GeoCLR with a SVM classifier (C2) outperforms all other configurations except for B2 when $M{=}40$.
     The best performance for $M{=}40$ is achieved by GeoCLR with a linear classifier (C1).
     (b) In general, the use of \emph{H-$k$means} improves performance compared to equivalent \emph{random} configurations, and GeoCLR outperform their counterparts with the SimCLR except for \{$M{=}40$, E4\}.
     }
\end{figure}

Figure~\ref{fig. acc plot random hk} shows representative configurations in Table~\ref{tab. data selection comparison}.
The configurations with the GeoCLR (F$\ast$) outperform their counterparts with the SimCLR (E$\ast$) except for the case where $M{=}40$ where E4 performs better than F4. In general, the use of \emph{H-$k$means} improves performance compared to equivalent \emph{random}
 configurations, achieving performance gains of 13.1\,\% and 5.7\,\% respectively for $M{=}40,100$. Although the gain in performance reduces for larger $M$, for GeoCLR \emph{H-$k$means} selection always improves performance compared to equivalent \emph{random} configurations for all values of $M$. An important observation is that the proposed GeoCLR achieved the best performance for all values of $M$ for both the \emph{balanced} and \emph{H-$k$means} selection strategies. 

A comparison between Table~\ref{tab. cnn training comparison} and Table~\ref{tab. data selection comparison} shows that GeoCLR with \emph{H-$k$means} performs better than with the \emph{balanced} selection strategy for all values of $M$, with gains of 3.1\,\% and 2.7\,\% for small values of $M{=}40,100$, and averaging a performance gain of 1.6\,\%. 

\begin{figure}[!ht]
    \centering
    \begin{subfigure}[b]{0.32\textwidth}
        \centering
        \includegraphics[width=\textwidth]{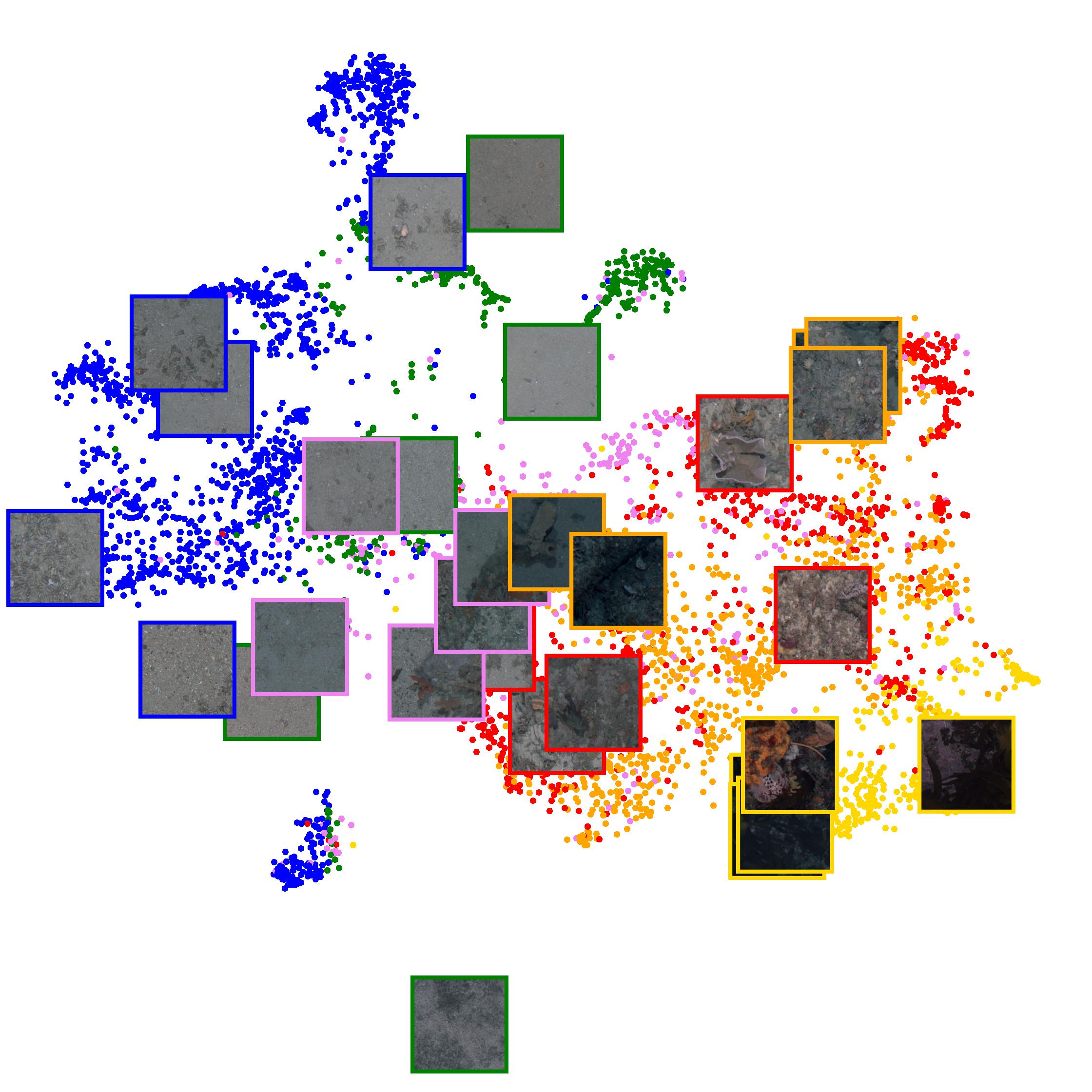} 
        \caption{\emph{Balanced}}
        \label{}
    \end{subfigure}
    \begin{subfigure}[b]{0.32\textwidth}
        \centering
        \includegraphics[width=\textwidth]{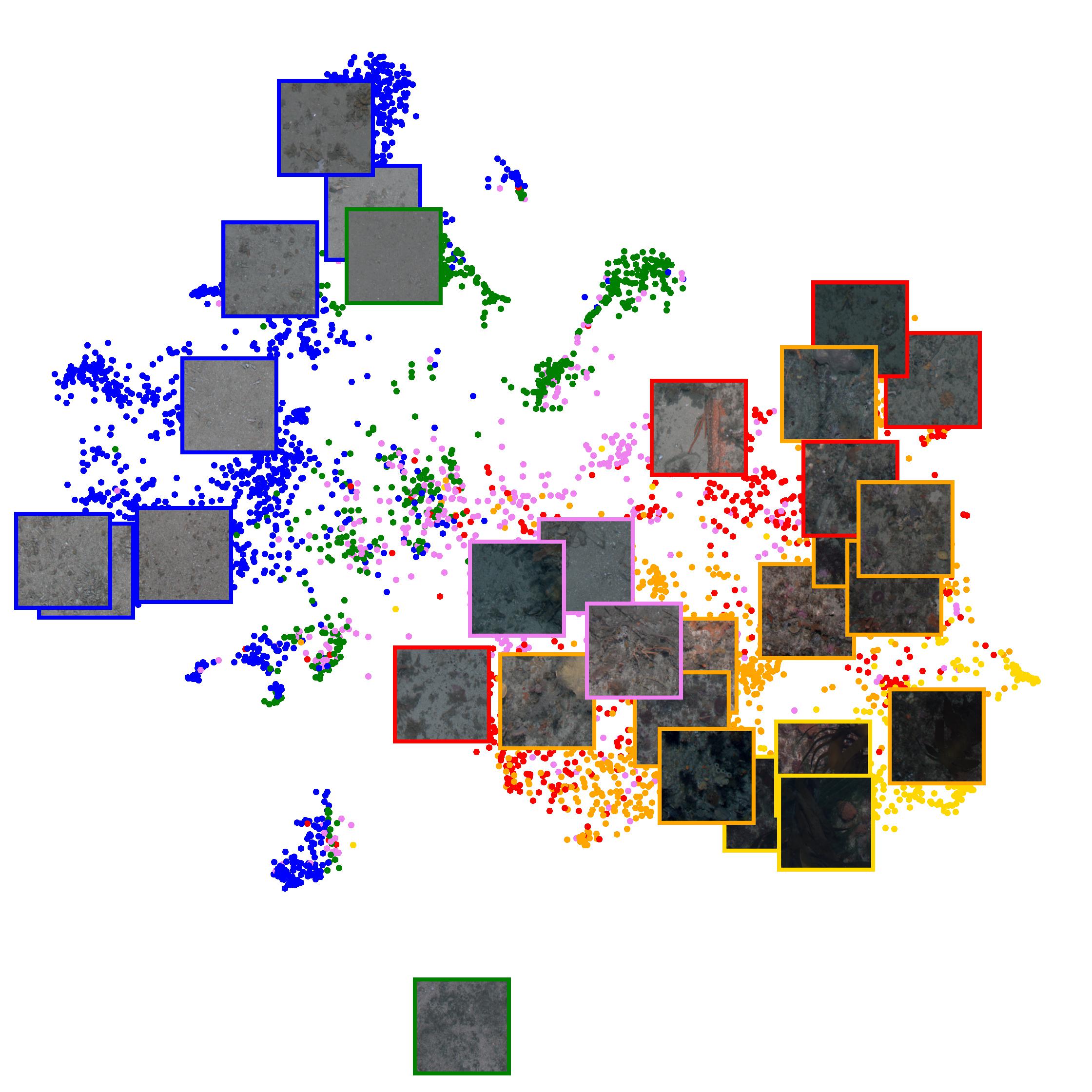} 
        \caption{\emph{Random}}
        \label{}
    \end{subfigure}
    \begin{subfigure}[b]{0.32\textwidth}
        \centering
        \includegraphics[width=\textwidth]{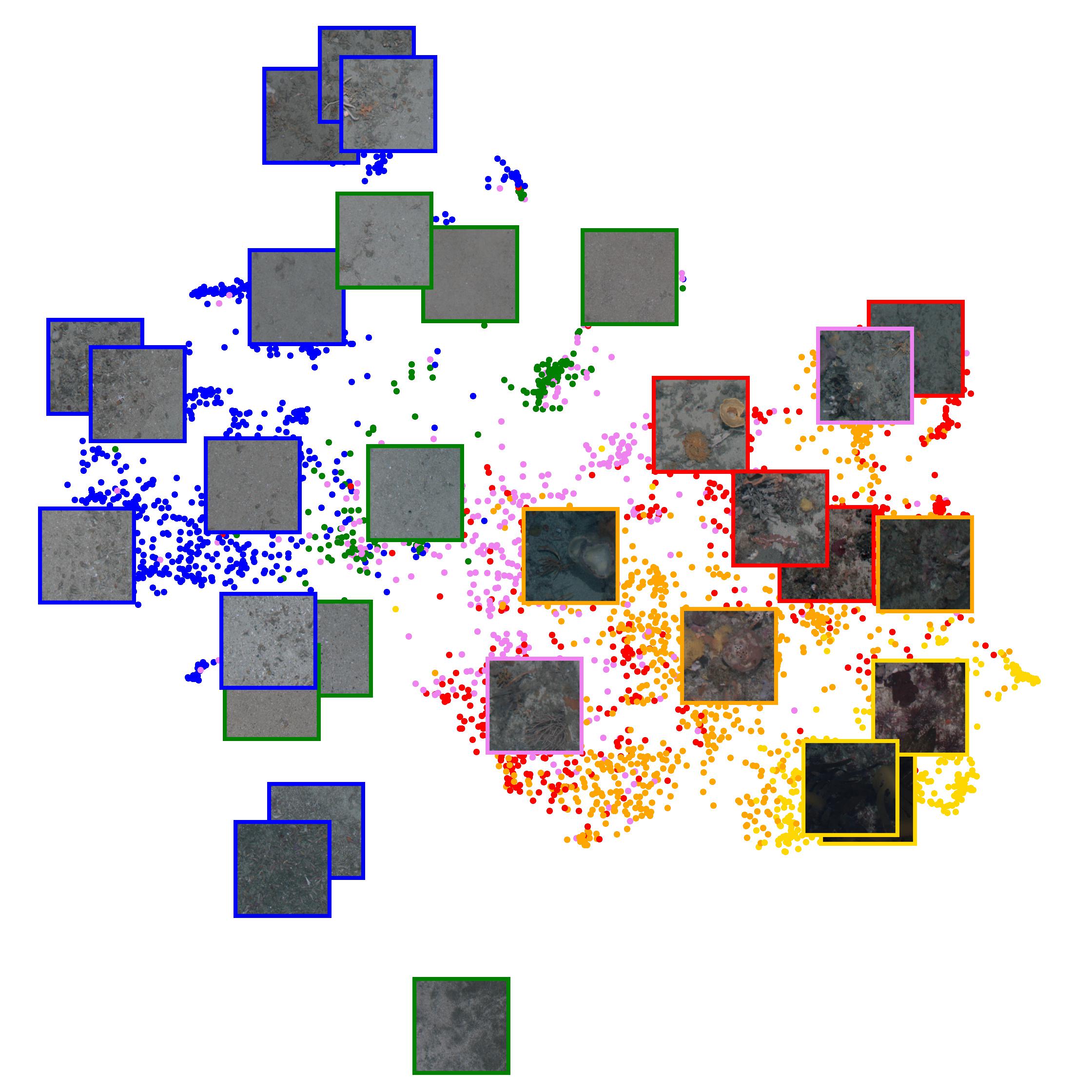} 
        \caption{\emph{H-kmeans}}
        \label{}
    \end{subfigure}
    \caption{
    Comparison of training data sampling strategy.
    $M=30$ images are selected by (a) \emph{Balanced}, (b) \emph{Random} and (c) \emph{H-kmeans} strategy.
    The selected images are shown on $t$-SNE visualisation of latent representations obtained by GeoCLR.
    While \emph{Random} sampling fails to select from the centre area, \emph{H-kmeans} successfully select the images in the relatively sparse areas so that a more informative training dataset is gained. Similarly, \emph{balanced} fails to sample regions of the latent space where there are more densely populated regions of the same class. In these situations, class assigned by the classifier will depend on the class of  training examples that happen to be nearby.
    The same colour scheme as Figure~\ref{fig. class example} is applied.
    }
    \label{fig. tsne}
\end{figure}

The results indicates that it is more valuable and informative to provide training data that evenly describes the latent representation space generated during self-supervised training, than it is to provide training data that evenly describes the targets that are of final interest to humans. Figure~\ref{fig. tsne} shows the representative images selected by (a) \emph{Balanced}, (b) \emph{Random} and (c) \emph{H-kmeans} strategies based on their location in the GeoCLR latent representations embedded by $t$-SNE\cite{maaten2008visualizing}. In this figure, $M{=}30$ images are shown for ease of visualisation, where the background points show the image representations that are not selected. The colour of the points and selected image borders illustrate the human class annotation of each annotated image using the same colour key as Figure \ref{fig. class example}. The visualisation shows that \emph{random} selection strategy fails to select images from the central region of the latent representation space that is relatively sparsely populated. On the other hand, \emph{H-kmeans} selects images evenly from the different regions of the latent representation. When compared to the \emph{balanced} selection strategy, it can be seen that there are several regions of the latent representation space that are not sampled. This is because they are mapped to different regions of the latent space as more densely populated regions that have the same class. These undersampled regions of the latent space can be easily confused by a classifier, where the final assigned class will depend on the distribution of nearby training samples. 

From a practical perspective, the proposed GeoCLR with $M{=}100$ \emph{H-$k$means} machine prioritised annotations and the SVM-RBF classifier (F4), and PL-SVM fine tuning (F7) achieves the same accuracy as state-of-the-art transfer learning (i.e. D5 $M{=}1000$) using an order of magnitude fewer human annotations. The method also achieves the same accuracy as state-of-the-art contrastive learning approaches (i.e. E3 $M{=}400$) using a quarter of the annotations, where prior works rely on random data annotations and do not propose a data selection strategy. We consider being able to perform accurate classification with a relatively small number of labels (i.e. 0.1\,\% of the entire dataset) an important development since providing 100 annotations represents a level of human effort that can be justified for most application in the field. We also show that for applications that can justify a larger amount of human effort (i.e. $M=1000$), the proposed GeoCLR outperforms conventional transfer learning (D5) and contrastive learning (E3) by 8.5\,\% and 7.5\,\% respectively. In addition to the demonstrated performance gains, the use of GeoCLR consistently improves performance over alternative configurations for all conditions tested in this work, and machine guided annotation \emph{H-$k$means} benefits performance for all configurations where $M{<}400$, and although the performance gains diminish for larger $M$, it never leads to significant performance reduction. The results indicate that these approaches can robustly improve the performance of CNNs for seafloor image interpretation.

\newpage

\subsection{Applications}
Determining seafloor habitat class distributions is a fundamental task for marine monitoring and conservation. Here we apply the proposed GeoCLR method to estimate the relative proportion of habitat classes and map their physical distribution.

\subsubsection{Estimating relative habitat class proportion}
\begin{figure}[!ht]
     \centering
     \begin{subfigure}[b]{0.80\textwidth}
         \centering
         \includegraphics[width=\textwidth]{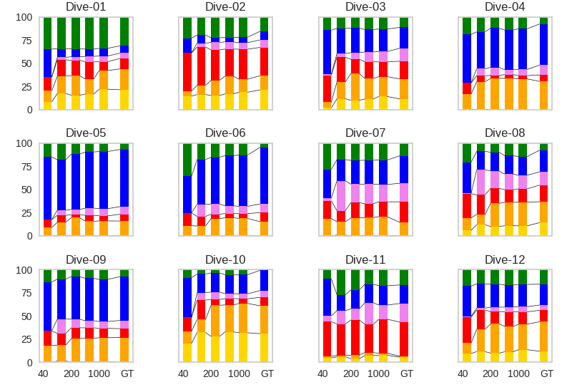}
         \caption{F4 in Table~\ref{tab. cnn training comparison} (SVM on latent representation $\boldsymbol{h}$)}
         \label{fig. proportion svm}
     \end{subfigure}\\
     \vspace{5 mm}
     \centering
     \begin{subfigure}[b]{0.80\textwidth}
         \centering
         \includegraphics[width=\textwidth]{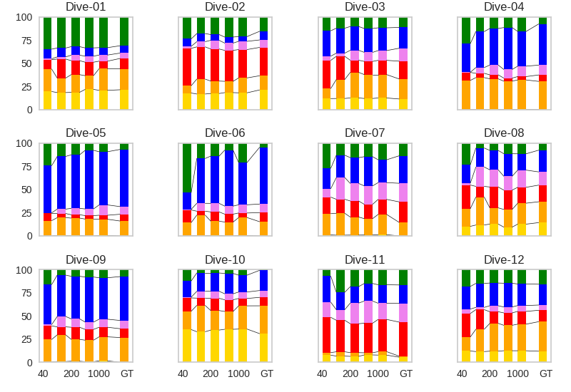}
         \caption{F8 in Table~\ref{tab. cnn training comparison} (fine-tuned on pseudo-labels generated by SVM)}
         \label{fig. proportion ft}
     \end{subfigure}
     \caption{Class distribution estimated for each dive based on the proposed GeoCLR.
     The same colour scheme is used as in Figure~\ref{fig. class example}.
     The estimated distributions approaches the ground truths when larger number of annotations are used for classifier training. The use of pseudo-labels is generally favourable for a small number of annotations (i.e. $M{=}40$, though this is not always the case. For $M{>}100$, F4 performs better than F8 and provides more stable estimates of class distribution as the same latent representation space is used for all $M$.}
     \label{fig. proportion}
\end{figure}

Figure~\ref{fig. proportion} shows the relative proportion of different habitat classes estimated for $M{=}[40,100,200,400,1000]$ machine prioritised annotations for each of the 12 dives in the Tasmania dataset. These are compared to the relative proportions for each dive where all human annotations have been used (i.e. average 450 annotations per dive) which we consider to be the ground truth here. The equivalent number of annotations per dive for the proposed method average approximately 3 annotations per dive for $M{=}40$ to approximately 83 per dive for $M{=}1000$. The results show that the estimated proportions approach the ground truth distributions for all dives, with the expected result that performance increases as a larger number of annotations are used for classifier training. The estimated proportions are poor for several of the dives with $M{=}40$ when using the F4 SVM classifier (Figure~\ref{fig. proportion svm}), whereas the F8 fine-tuned with pseudo-labels generated by the SVM is generally more robust, approximating the ground truth class proportions better for the same number of training examples (Figure~\ref{fig. proportion ft}). This indicates that the SVM classifier (F4) may be overfitting the latent representation space generated by GeoCLR when the number of annotations available is small, where this effect is mitigated by providing a larger number of training examples through pseudo-labels. However, there are some exception (Dives 06 and 07) to this and so the outputs with $M{=}40$ should be treated with caution where validation data is not available. On the other hand, for $M{\ge}100$ both methods (i.e. F4 and F8) perform robustly for all dives, with F4 outperforming F8 and providing more stable estimates for different values of $M$. This is due to the fact that the latent representation space remains the same regardless of $M$ as no CNN re-training takes place. 

\subsubsection{Habitat mapping}
The physical distribution of habitats is important for conservation since it influences the distribution of organisms near the seafloor. It is also important for understanding ecosystem health as benthic habitats such as kelp (seen here) and coral are classified as essential ocean variables.

\begin{figure}[!ht]
    \centering
    \begin{subfigure}[b]{0.58\textwidth}
        \centering
        \includegraphics[width=\textwidth]{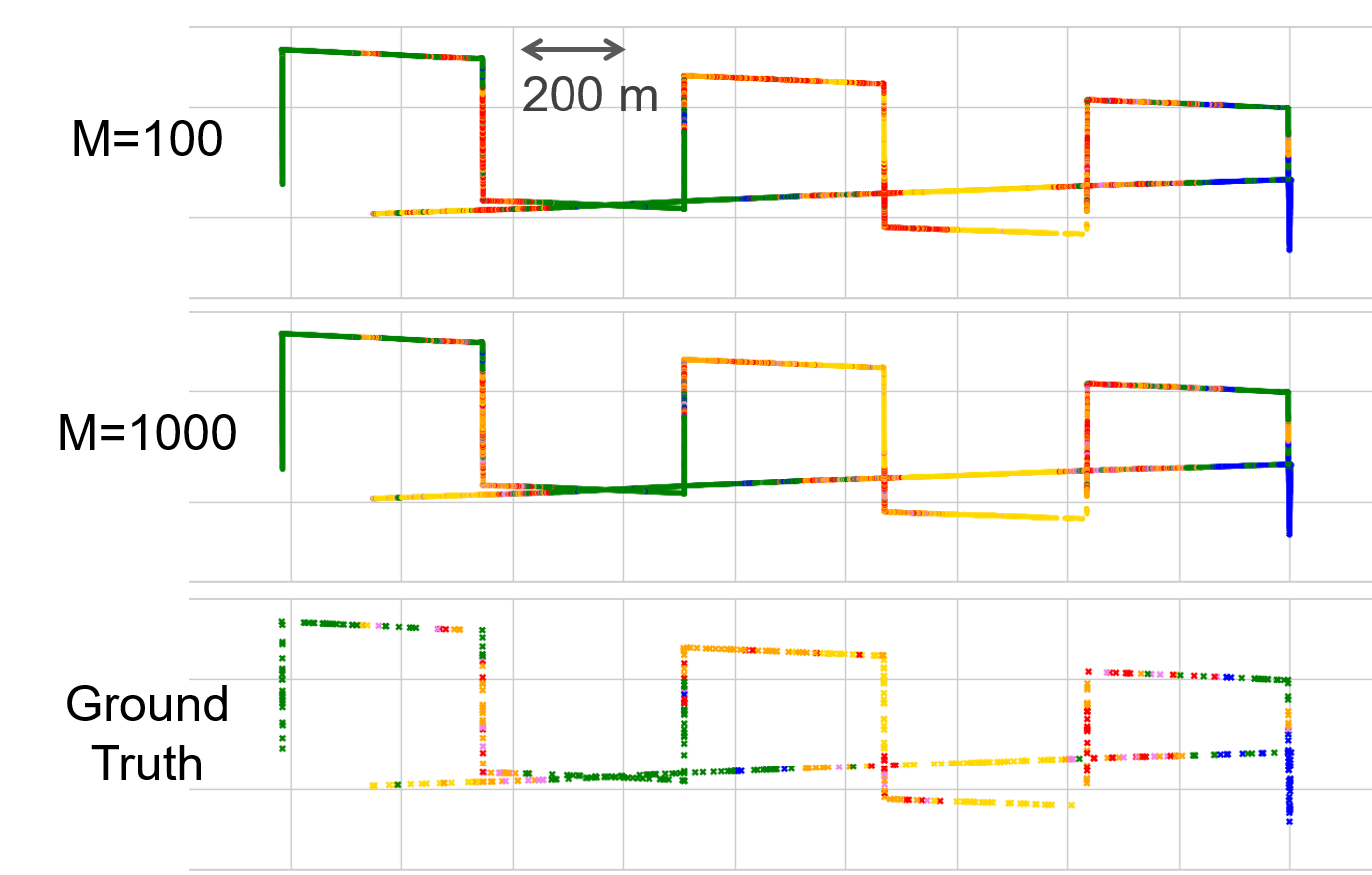} 
        \caption{Horizontal distribution}
        \label{}
    \end{subfigure}
    \begin{subfigure}[b]{0.38\textwidth}
        \centering
        \includegraphics[width=\textwidth]{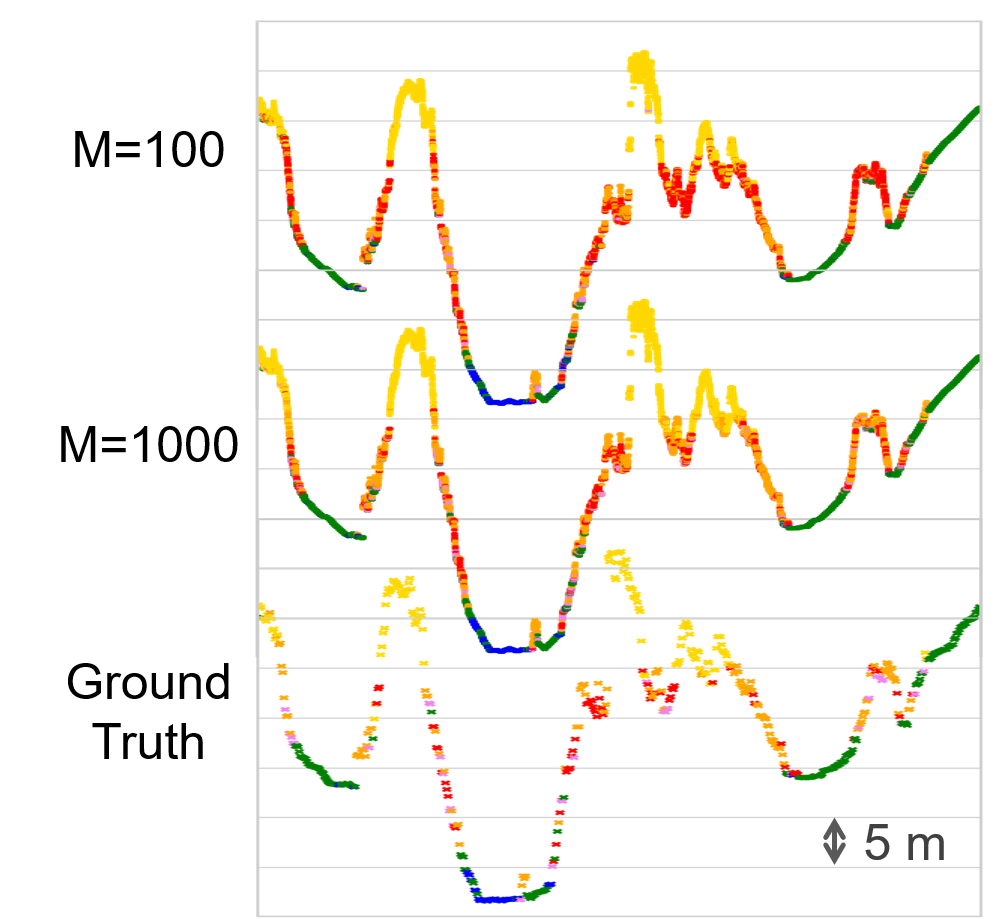} 
        \caption{Depth profile vs. image index}
        \label{}
    \end{subfigure}\\
    \vspace{10 mm}
    \begin{subfigure}[b]{0.9\textwidth}
        \includegraphics[width=1\textwidth]{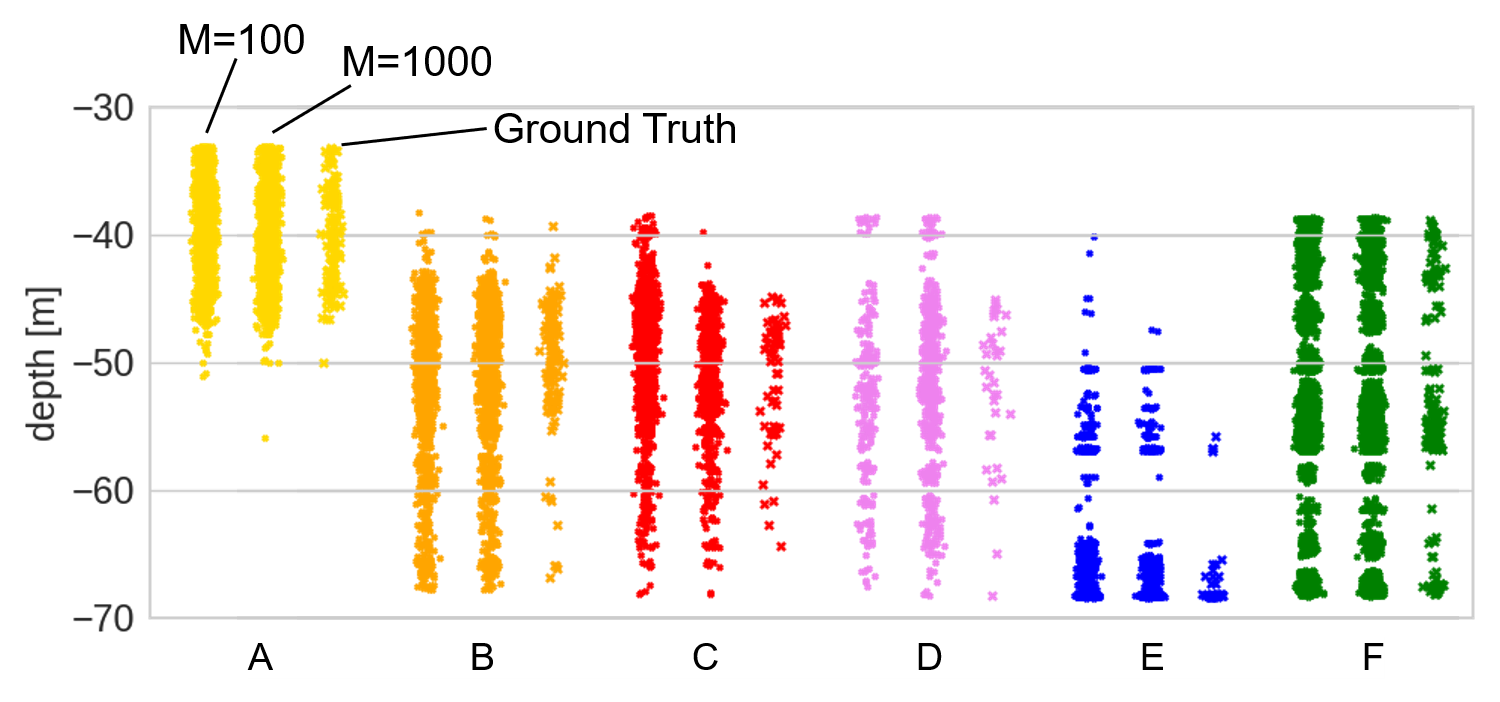} 
        \caption{Class vs. Depth}
        \label{}
    \end{subfigure}
    \caption{Class distribution of Dive-01 with $M${=}100 annotations by F8, $M${=}1000 annotations by F4 in Table~\ref{tab. data selection comparison} and ground truth.
    The same colour scheme as Figure~\ref{fig. class example} is applied.}
    \label{fig. dive01}
\end{figure}

\begin{figure}[!ht]
    \centering
    \begin{subfigure}[b]{0.58\textwidth}
        \centering
        \includegraphics[width=\textwidth]{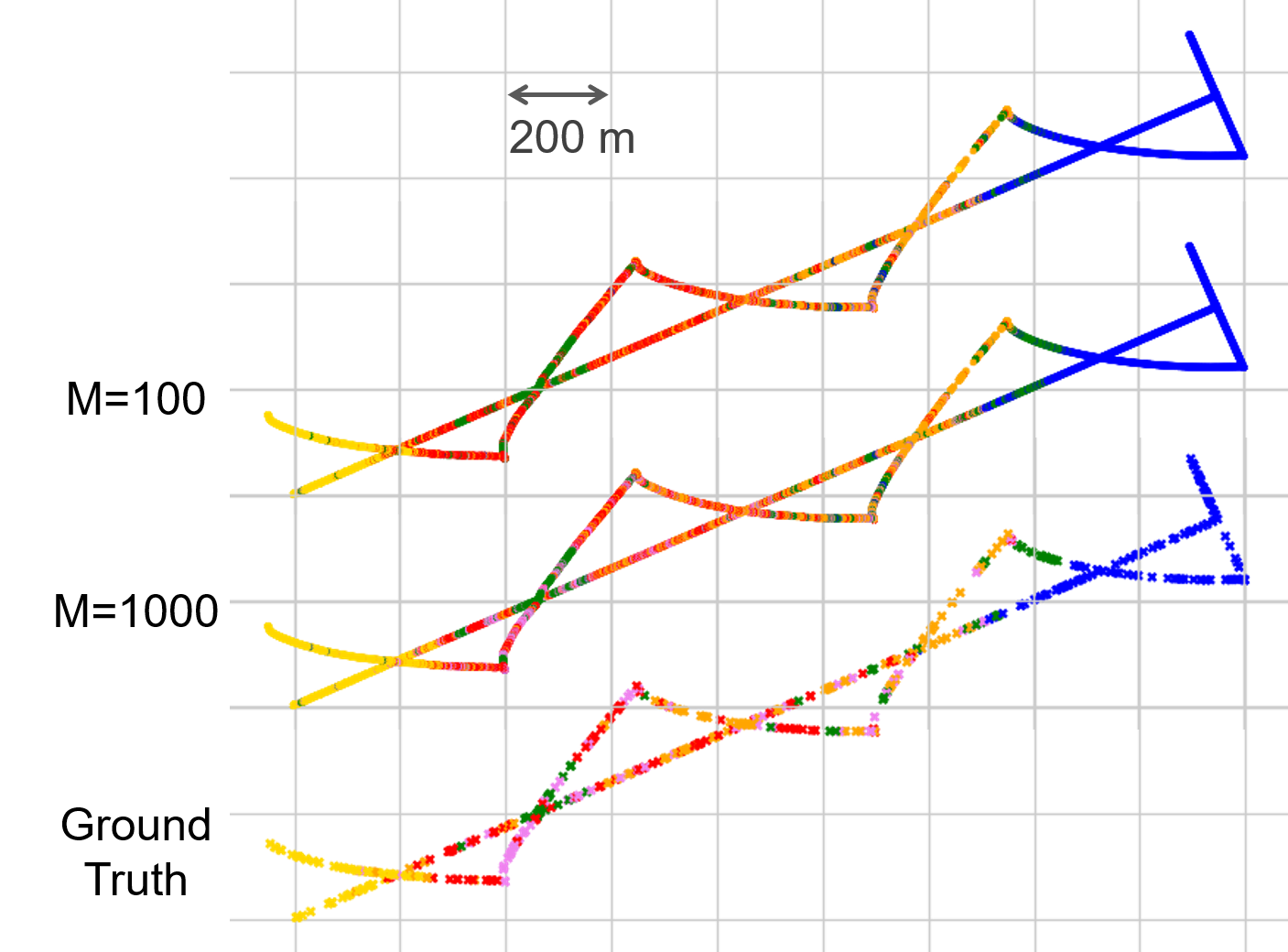} 
        \caption{Horizontal distribution}
        \label{}
    \end{subfigure}
    \begin{subfigure}[b]{0.38\textwidth}
        \centering
        \includegraphics[width=\textwidth]{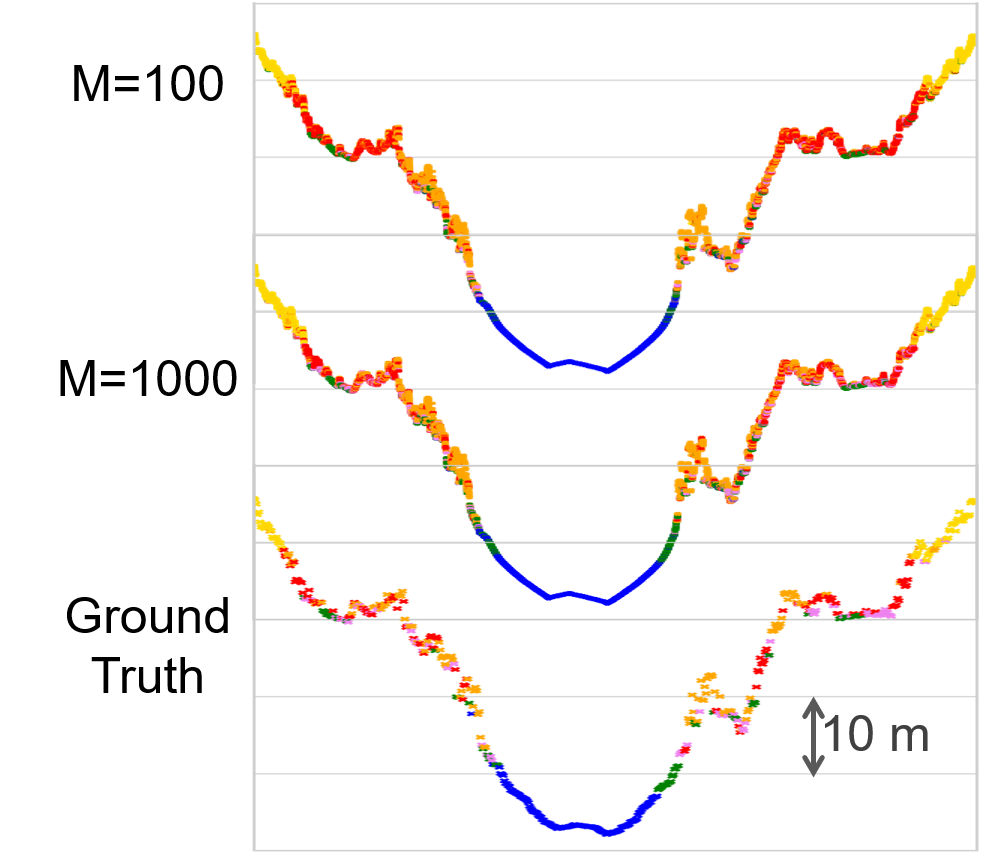} 
        \caption{Depth profile vs. image index}
        \label{}
    \end{subfigure}\\
    \vspace{10 mm}
    \begin{subfigure}[b]{0.9\textwidth}
        \includegraphics[width=1\textwidth]{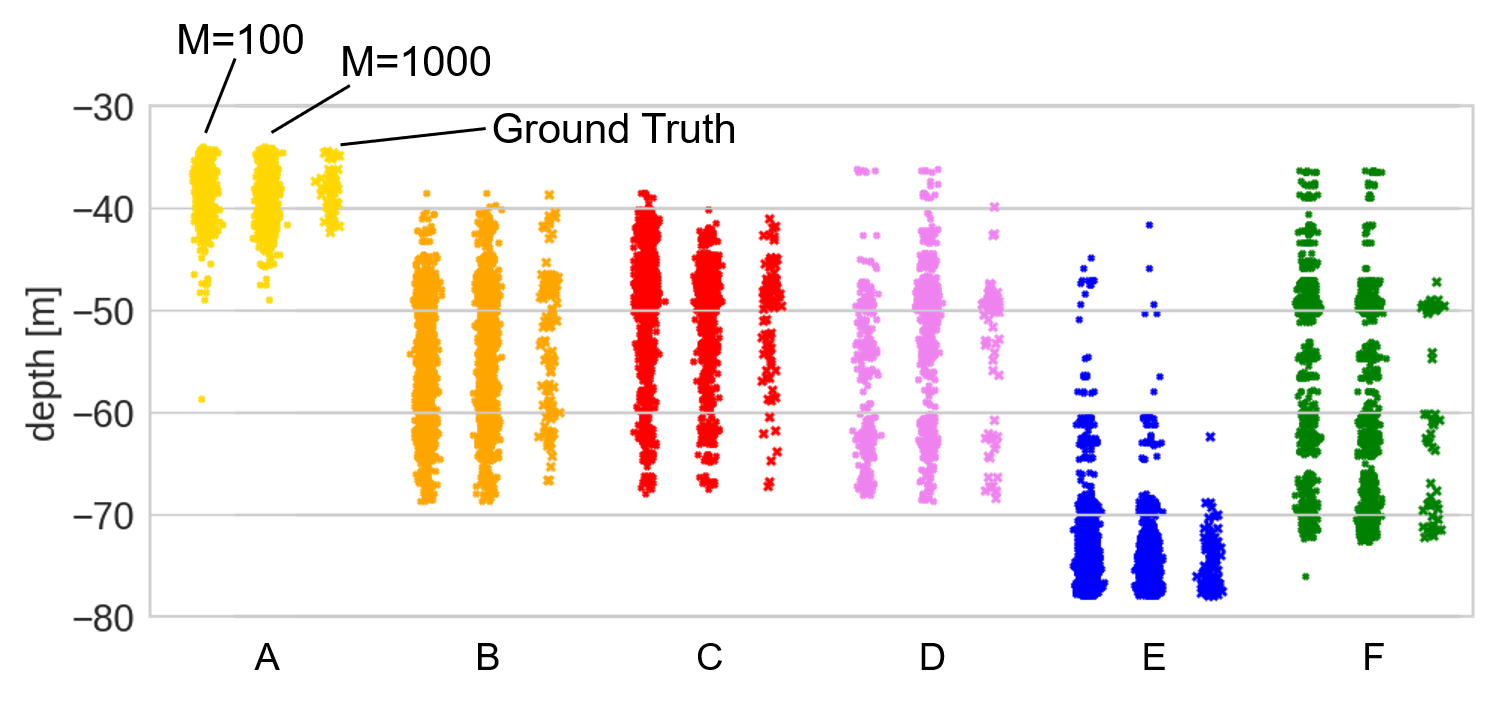} 
        \caption{Class vs. Depth}
        \label{}
    \end{subfigure}
    \caption{Class distribution of Dive-03 with $M${=}100 annotations by F8, $M${=}1000 annotations by F4 in Table~\ref{tab. data selection comparison} and ground truth.
    The same colour scheme as Figure~\ref{fig. class example} is applied.}
    \label{fig. dive03}
\end{figure}

\begin{figure}[!ht]
    \centering
    \begin{subfigure}[b]{0.58\textwidth}
        \centering
        \includegraphics[width=\textwidth]{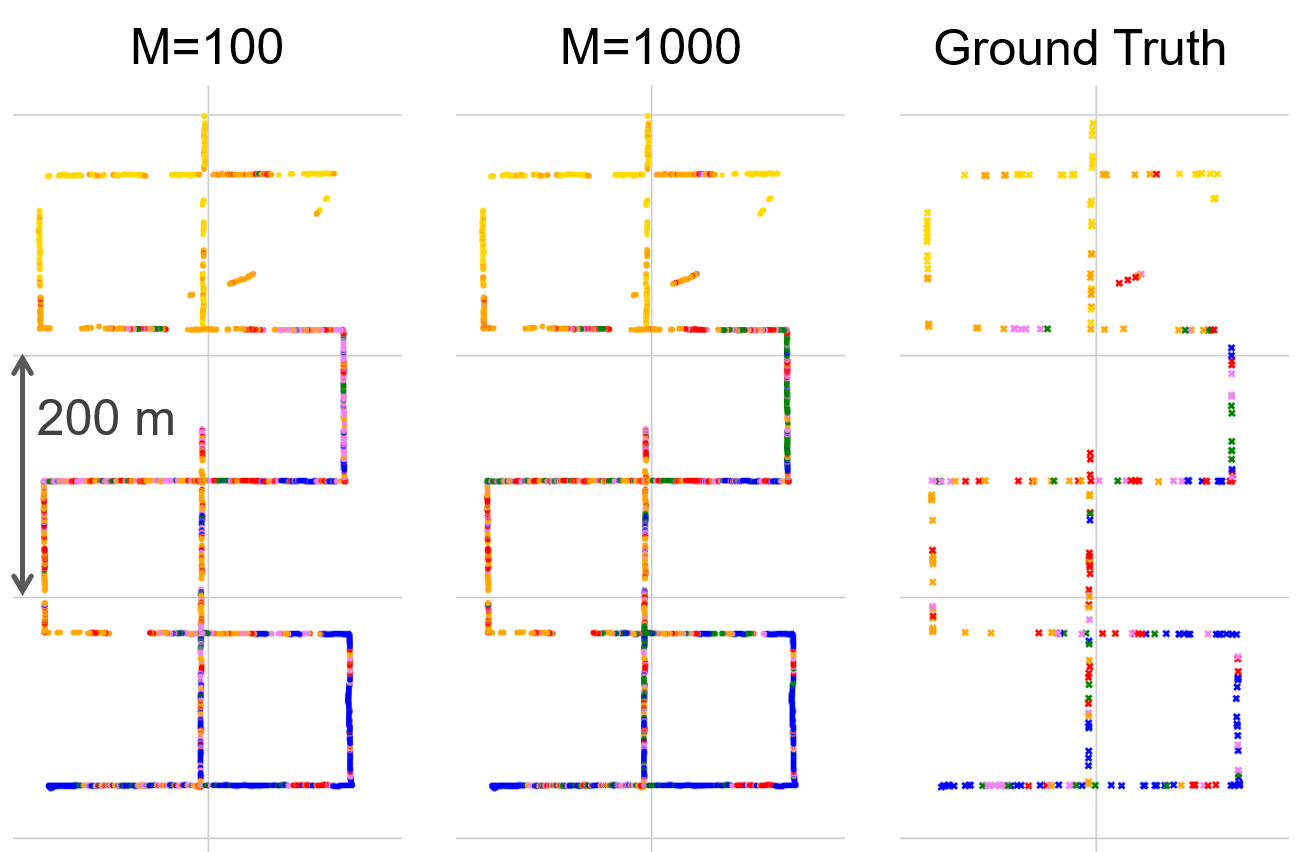} 
        \caption{Horizontal distribution}
        \label{}
    \end{subfigure}
    \begin{subfigure}[b]{0.38\textwidth}
        \centering
        \includegraphics[width=\textwidth]{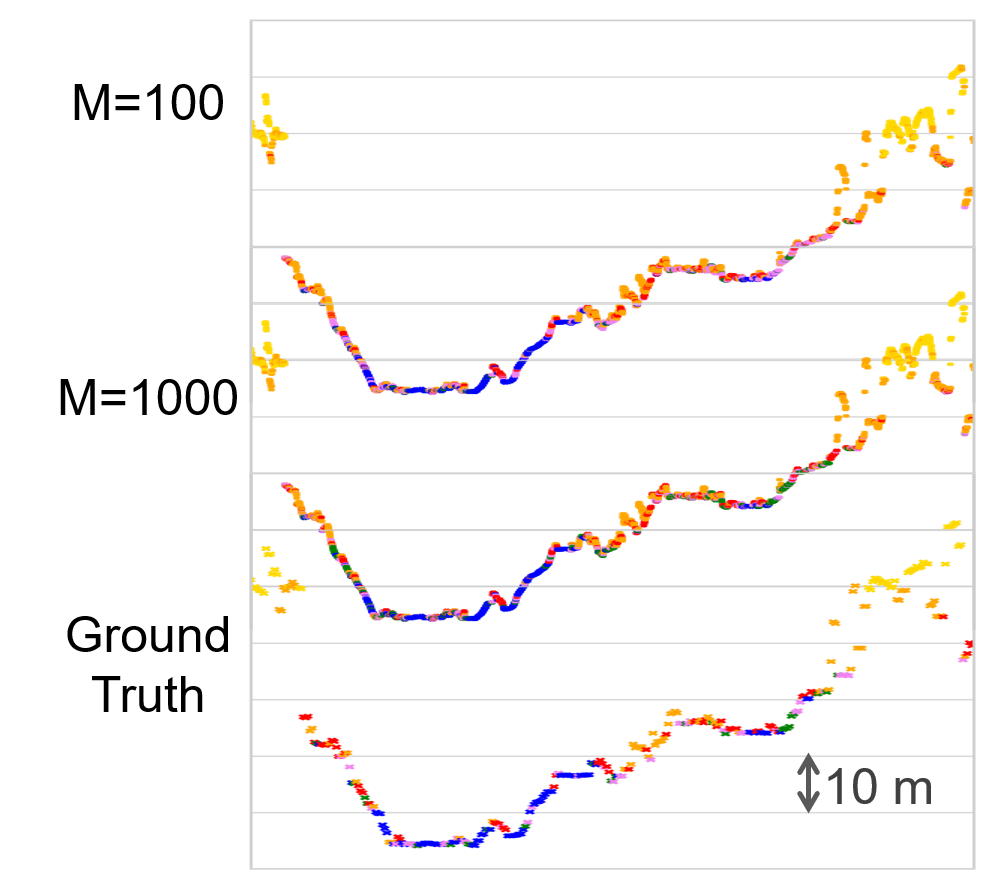} 
        \caption{Depth profile vs. image index}
        \label{}
    \end{subfigure}\\
    \vspace{10 mm}
    \begin{subfigure}[b]{0.9\textwidth}
        \includegraphics[width=1\textwidth]{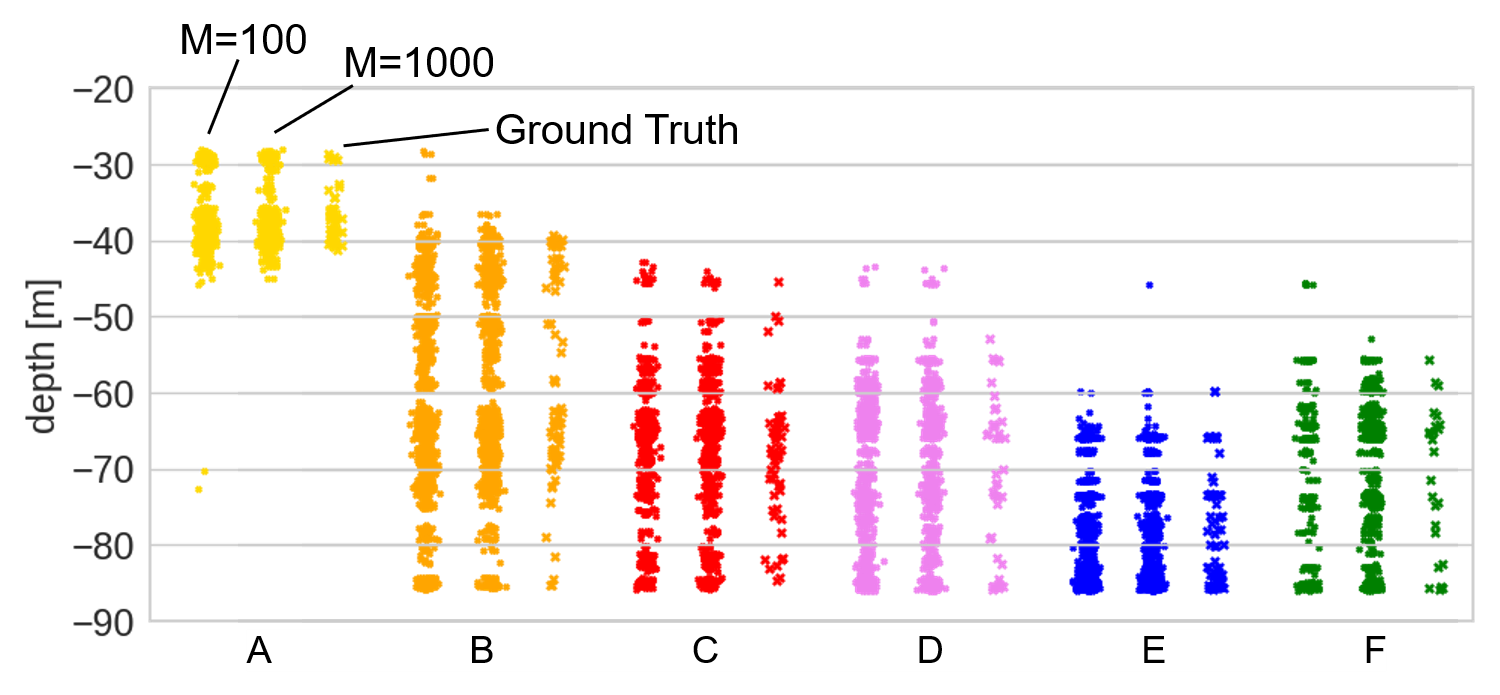} 
        \caption{Class vs. Depth}
        \label{}
    \end{subfigure}
    \caption{Class distribution of Dive-08 with $M${=}100 annotations by F8, $M${=}1000 annotations by F4 in Table~\ref{tab. data selection comparison} and ground truth.
    The same colour scheme as Figure~\ref{fig. class example} is applied.}
    \label{fig. dive08}
\end{figure}

The proposed method allows efficient estimation of habitat maps based on the 3D location where each classified image was taken. Here, we show the horizontal distributions of the classes, the depth profiles vs. image index, and the class vs. depth distributions are shown in Figures~\ref{fig. dive01},~\ref{fig. dive03} and ~\ref{fig. dive08} for three dives (01,03 and 08) which were chosen as representative cases. The figures show habitat maps generated using GeoCLR for \{$M{=}100$,F8\}, \{$M{=}1000$,F4\} in Table~\ref{tab. data selection comparison} and the ground truth labels.

The results show that both \{$M{=}100$,F8\} and \{$M{=}1000$,F4\} configurations closely approximate the ground truth horizontal and vertical habitat class distributions, capturing the continuous spatial transitions between Kelp (A), Low Relief Reef (C), High Relief Reef (B) to Screw Shell Rubble (E) or Sand (F). The class vs depth distributions show that the larger values of $M$ provide a better approximation of vertical class distribution, which is an expected result. However, for classes that exist in a limited depth band (e.g. Kelp (A), Screw Shell Rubble (E)) both values of $M$ capture this trend.

\clearpage
\section{Conclusion}
The paper has developed a method to leverage georeferenced information in contrastive learning for efficient training of deep-learning CNNs. The proposed Georeference Contrastive Learning for seafloor image Representation (GeoCLR) method is effective for datasets where the targets of interest for semantic interpretation are distributed over spatial scales larger that the footprint of a single image frame. The underlying assumption is that images gathered within a close distance are more likely to have similar visual appearance than images that are taken far apart. This assumption can be reasonably satisfied in seafloor robotic imaging applications, where 1) images are acquired at close and regular intervals along a vehicle's trajectory, and 2) the targets for interpretation are substrates and habitats which typically have extents much greater than the image footprint. The method can be deployed on any CNN, and performance gains can be achieved without any prior human input to interpret the dataset. We demonstrate the performance of the proposed training method using a CNN architecture that can be deployed on computers that can be reasonably expected to be available during a field survey, without relying on network access to supercomputers, and can generate results in timeframes that are relevant for on-going field expeditions. Experiments on a robotically obtained seafloor image dataset that includes more than \url{~}86k images and \url{~}5k annotations show that:

\begin{itemize}
    \item The proposed GeoCLR method outperforms existing state-of-the-art contrastive learning (SimCLR) and transfer learning for downstream supervised classification tasks using an equivalent CNN architecture (ResNet18). On an ideal, class balanced training dataset, the SVM with RBF kernel trained on the features extracted by the GeoCLR trained CNN shows an average of 5.2 \% and maximum of 7.7 \% improvement compared to the accuracy scores of SimCLR for $M{=}[40,100,200,400,1000]$ annotations. %
    Compared to ResNet18 trained by transfer learning, an average improvement of 7.4 \%, a maximum of 9.2 \% is achieved. %
    \item The representations extracted by the GeoCLR are useful for identifying representative images for prioritised human annotation in a fully unsupervised manner. This can improve the performance and efficiency of human effort for classification, where selecting a prioritised training dataset using H-$k$ means clustering increases the classification accuracy by an average of 4.9 \% and maximum of 14.1 \% compared to random selection, where the performance gains are more significant for small numbers of $M$.
    Compared with SimCLR, GeoCLR latent representations shows 7.1 \% better score on average with randomly annotated $M{=}[40,100,200,400,1000]$ training datasets.
    Prioritised annotating by \emph{H-$k$means} allows score improvements for all $M$ values, leading to 10.2 \% increase in total. %
    
    \item Selecting representative images for prioritised labelling based on their distribution in the GeoCLR latent representation space results in better performance than providing class-balanced annotated examples. The machine driven \emph{H-$k$means} selection strategy achieves an average of 1.6\,\% and maximum of 3.1\,\% increase in accuracy compared to the class-balanced selection strategy for an equivalent number of annotations, where greater gains are achieved for small numbers of annotations. This indicates that it is more informative to provide training data that evenly describes the latent representation space generated during self-supervised training, than it is to provide training data that evenly describes the targets that are of final interest to humans.
    \item The combination of GeoCLR and \emph{H-$k$means} achieves the same accuracy as state-of-the-art transfer learning using an order of magnitude fewer human annotations, and state-of-the-art contrastive learning approaches using a quarter of the labels. This allows the proportion of habitat classes and their spatial distribution to be accurately estimated ($>70\,\%$) annotating only 0.1 \% of the images in the dataset. This is significant as providing approximately 100 annotations represents a level of human effort that can be justified for most field application. For applications where a greater level of human effort is available, we show that with 1000 annotations, the proposed GeoCLR outperforms conventional transfer learning and contrastive learning by 8.5\,\% and 7.5\,\% respectively, achieving a classification accuracy of $79\,\%$. The combination of GeoCLR and  \emph{H-$k$means} never degraded performance compared to equivalent alternative configurations in the experiments described in this paper.
\end{itemize}

\subsubsection*{Acknowledgements}
This work was carried out under the UK Natural Environment Research Council’s Oceanids Biocam project NE/P020887/1 and Australian Research Council’s Automated Benthic Understanding DP190103914 Discovery project.

\newpage
\bibliographystyle{apalike}
\bibliography{contrastive_learning_refs}

\newpage
\appendix

\setcounter{table}{0}
\renewcommand{\thetable}{A\arabic{table}}
\setcounter{figure}{0}
\renewcommand{\thefigure}{A\arabic{figure}}

\section{Sensitivity to hyperparameters}\label{Sec. Sensitivity of hyperparameters}
The main contribution of the GeoCLR method is that selecting similar image pairs that are physically close to each other will provide a better representation of variability in contrastive learning than traditional data augmentation. 

The conditions for a pair of images to be physically close enough is determined by the following equation (eq. (3) in the main text):

\setcounter{equation}{2}
\begin{equation} \label{eq. distance criteria}
	\sqrt{(g_{east}'-g_{east})^2 + (g_{north}'-g_{north})^2 + \lambda (g_{depth}'-g_{depth})^2 } \leq r,
\end{equation}

where $(g_{east},g_{north},g_{depth})$ is the 3D georeference of image $\boldsymbol{x}$.
Eq. (3) includes two hyperparameters: $r$, which corresponds to a 3D distance threshold where images captured within this range are regarded as similar, and $\lambda$, which is a weight for the depth difference between images.
These hyperparameters relate to the physical characteristics of observed seafloor habitats and substrates, and so their optimised values are considered to be dataset dependent.
This appendix investigates the sensitivity of learning performance to these hyperparameters for the Tasmania dataset considered in this work.

\begin{table}[h]
	\centering
	\caption{Performance sensitivity to hyperparameter $r$ (in metres) when validated on class balanced training subsets}
	\label{tab. r}
	\centering
	
	\begin{tabular}{c|cccccc}
		\hline \hline
		\multirow{2}{*}{\textbf{\begin{tabular}[c]{@{}c@{}}\\  $r$\end{tabular}}} &
		\multicolumn{5}{c}{\textbf{Number of Annotations ($M$)}} \\
		& 40 & 100 & 200 & 400 & 1000            \\
		
		\hline
		0.0 (SimCLR)& 62.5$\pm$2.7 & 65.2$\pm$2.8 & 67.1$\pm$1.2 & 69.2$\pm$2.2 & 71.8$\pm$1.0\\
		1.0 & 63.8$\pm$2.9 & 67.8$\pm$2.4 & 71.4$\pm$1.4 & 72.9$\pm$1.8 & 74.9$\pm$1.0\\
		3.0 & 59.7$\pm$2.7 & 64.4$\pm$2.4 & 69.2$\pm$3.4 & 70.8$\pm$2.1 & 72.8$\pm$1.4\\
		5.0 & 61.1$\pm$2.2 & 65.9$\pm$2.4 & 68.4$\pm$1.6 & 71.1$\pm$1.8 & 71.9$\pm$2.3\\
		10.0 & 60.5$\pm$2.5 & 66.1$\pm$3.0 & 68.5$\pm$2.5 & 70.5$\pm$3.1 & 72.9$\pm$1.7\\
		
		\hline \hline
	\end{tabular}
	\begin{flushleft}
		$r=0.0$ is equivalent to SimCLR (B1 in Table 2).
		
	\end{flushleft}
\end{table}

Table \ref{tab. r} shows the f$_1$ scores of linear classifiers trained on the obtained latent representations for different  $r$ values ($r {=} 0.0, 1.0, 3.0, 5.0, 10.0$\,m) for fixed $\lambda {=} 1.0$.
Class balanced training datasets with different numbers of images ($M {=} 40, 100, 200, 400, 1000$) have been used to train the classifier, which is the same configuration as B1 (for $r {=} 0.0$\,m) or C1 (others) in Table 2 of the main paper.
The best performance is using $r {=} 1.0$\,m for all $M$, showing that it is optimal for this dataset.
Since the AUV travelled at $0.5 m/s$ with image acquisition at 1 fps, $r {=} 1.0$\,m is the smallest value where most images will have at least two nearby images to form a similar pair for contrastive learning. Larger $r$ values will have more images to chose from for the similar pair, but as the distances between the pair increases, we expect their appearances to become less similar. If $r$ is too small, there will be no nearby images to select from, and so only augmentation on the same image can be performed (i.e., the same as SimCLR). The latent representations obtained with the larger $r$ values ($r {=} 3.0, 5.0, 10.0$\,m) show poorer performance scores than $r {=} 1.0$\,m for all $M$. 
However, they still perform better than the original SimCLR ($r {=} 0.0$\,m) except for a few cases with $M{=}40,100$.
It can be assumed that the optimal $r$ value depends on the habitat and substrate patch sizes in the observed area. Even if the size of these semantic patches is not known, we can expect smaller $r$ values to perform robustly since similar appearance image pairs are likely to be sampled, compared with larger $r$, under the assumption that semantic patches of interest for habitat and substrate mapping occur on spatial scales larger than the footprint of a single image frame.

\begin{table}[h]
	\centering
	\caption{Performance sensitivity to hyperparameter $\lambda$ when validated on class balanced training subsets}
	\label{tab. lambda}
	\centering
	
	\begin{tabular}{c|cccccc}
		\hline \hline
		\multirow{2}{*}{\textbf{\begin{tabular}[c]{@{}c@{}}\\  $\lambda$\end{tabular}}} &
		\multicolumn{5}{c}{\textbf{Number of Annotations ($M$)}} \\
		& 40 & 100 & 200 & 400 & 1000            \\
		
		\hline
		0.0 & 63.8$\pm$3.3 & 69.2$\pm$2.9 & 72.5$\pm$1.7 & 74.0$\pm$1.9 & 75.6$\pm$1.3\\
		0.5 & 63.7$\pm$4.7 & 68.4$\pm$3.1 & 71.8$\pm$2.0 & 74.0$\pm$1.9 & 76.6$\pm$2.1\\
		1.0 & 63.8$\pm$2.9 & 67.8$\pm$2.4 & 71.4$\pm$1.4 & 72.9$\pm$1.8 & 74.9$\pm$1.0\\
		3.0 & 65.3$\pm$4.6 & 70.2$\pm$2.9 & 72.7$\pm$2.9 & 74.4$\pm$1.7 & 75.9$\pm$1.3\\
		5.0 & 63.8$\pm$2.6 & 69.6$\pm$3.4 & 72.3$\pm$2.3 & 74.2$\pm$1.7 & 75.3$\pm$1.7\\
		10.0 & 64.0$\pm$3.7 & 68.7$\pm$3.3 & 72.9$\pm$2.0 & 74.3$\pm$2.5 & 76.0$\pm$1.6\\
		
		\hline \hline
	\end{tabular}
	\begin{flushleft}
		
	\end{flushleft}
\end{table}

Table \ref{tab. lambda} shows the f$_1$ scores trained latent representations obtained using the optimal $r$ value ($r{=}1.0$\,m) with the different $\lambda$ values ($\lambda{=} 0.0, 0.5, 3.0, 5.0, 10.0$) for the Tasmania dataset. Increasing $\lambda$ increases sensitivity to depth differences, making it more likely for a potential similar pair to be rejected if there is a difference in their depths. Though the scores differ slightly depending on the $\lambda$ values, their standard deviation values show these differences within the margin of error.
The reason why $\lambda$ is less sensitive to the performance than $r$ can be considered that rugosity and slope in the Tasmania dataset are relatively small. For the datasets with more drastic depth changes we expect greater sensitivity to $\lambda$ values.

\end{document}